\documentclass[10pt,twocolumn,letterpaper]{article}

\usepackage[pagenumbers]{cvpr}     %

\usepackage{url}
\usepackage{amsfonts}
\usepackage{acronym}
\usepackage{multirow}
\usepackage{tikz}
\usetikzlibrary{backgrounds}
\usepackage{overpic}
\usepackage{tabularx}
\usepackage{makecell}
\usepackage{pifont}
\usepackage{colortbl}
\usepackage{float}
\usepackage{adjustbox}
\usepackage{comment}

\definecolor{groupbg}{gray}{0.92}
\definecolor{groupfg}{gray}{0.20}

\let\eg\undefined
\let\ie\undefined

\newdimen\typewidth
\AtBeginDocument{\typewidth=\textwidth}

\newlength\extramargin
\usepackage[disable]{todonotes}

\usepackage[
final,tracking=true,kerning=true,spacing=true,factor=1100,stretch=10,shrink=10]{microtype}
\microtypesetup{protrusion=false}

\definecolor{cvprblue}{rgb}{0.21,0.49,0.74}
\usepackage[pagebackref,breaklinks,colorlinks,allcolors=cvprblue]{hyperref}

\def\paperID{*****}
\def\confName{CVPR}
\def\confYear{2026}

\newcommand{\m}[1]{\mathcal{#1}}

\newcommand{\boldparagraph}[1]{\noindent\subsubsection{#1}}

\newcommand{\pcd}{point cloud\xspace}
\newcommand{\pcds}{point clouds\xspace}

\newcommand{\tdsgslong}{3D Scene Graphs\xspace}
\newcommand{\tdsglong}{3D Scene Graph\xspace}
\newcommand{\tdsgshort}{3DSG\xspace}
\newcommand{\tdsgsshort}{3DSGs\xspace}

\newcommand{\eg}{\emph{e.g.,}\xspace}
\newcommand{\ie}{\emph{i.e.,}\xspace}

\acrodef{SLAM}{Simultaneous Localization and Mapping}
\acrodef{VLA}{Vision Language Action model}
\acrodef{ATE}{absolute trajectory error}
\acrodef{RMSE}{Root Mean Squared Error}
\acrodef{GCN}{Graph Convolutional Network}
\acrodef{MLP}{Multi-Layer Perceptron}
\acrodef{LLM}{Large Language Model}
\acrodef{FM}{Foundation Model}
\acrodef{VLM}{Vision-Language Model}
\acrodef{MST}{Minimum Spanning Tree}
\acrodef{POV}{Point-of-View}
\acrodef{BEV}{bird’s-eye-view}
\acrodef{IMU}{inertial measurement unit}
\acrodef{VIO}{visual-inertial odometry}
\acrodef{SUN}{Scene Understanding}
\acrodef{XR}{extended reality}
\acrodef{TAMP}{task and motion planning}
\acrodef{HRI}{Human Robot Interaction}
\acrodef{POMDP}{partially observable Markov decision process}

\newcommand{\website}{\url{https://3dscenegraphs.com}}

\renewcommand{\boldparagraph}[1]{\noindent\textbf{#1.}}
\title{3D Scene Graphs: Open Challenges and Future Directions}

\author{%
Dennis Rotondi\textsuperscript{1,2,*}\quad
Francesco Argenziano\textsuperscript{3,*}\quad
Sebastian Koch\textsuperscript{4}\quad
Nathan Hughes\textsuperscript{5}\quad
Martin B\"uchner\textsuperscript{6}\\
Johanna Wald\textsuperscript{4}\quad
Lukas Rosenberger Schmid\textsuperscript{7}\quad
Daniele Nardi\textsuperscript{3}\quad
Abhinav Valada\textsuperscript{6}\quad
Liam Paull\textsuperscript{8,9}\\
Federico Tombari\textsuperscript{4,10}\quad
Luca Carlone\textsuperscript{5}\quad
Kai O. Arras\textsuperscript{1}\\[0.4em]
{\small\textsuperscript{1}University of Stuttgart\quad
\textsuperscript{2}IMPRS-IS\quad
\textsuperscript{3}Sapienza University of Rome\quad
\textsuperscript{4}Google\quad
\textsuperscript{5}MIT}\\
{\small\textsuperscript{6}University of Freiburg\quad
\textsuperscript{7}UTN\quad
\textsuperscript{8}University of Montreal\quad
\textsuperscript{9}Mila\quad
\textsuperscript{10}TU Munich}%
}

\begin{document}

\twocolumn[{%
\renewcommand\twocolumn[1][]{#1}%
\maketitle
\vspace{-3.0em}
\begin{center}
\includegraphics[width=\linewidth]{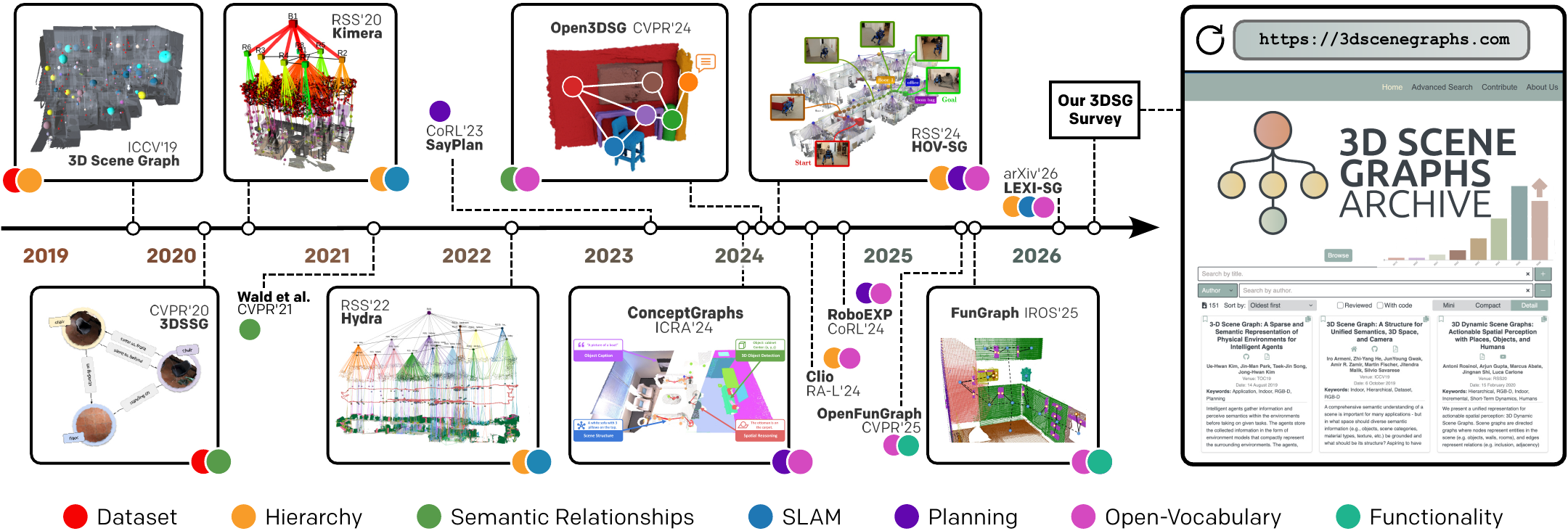}
    \captionof{figure}{\textbf{Overview of the evolution of 3D Scene Graph research (2019-2026).}
    Representative milestone papers are organized chronologically and color-coded by key research themes: datasets, hierarchy, semantic relationships, SLAM integration, planning, open-vocabulary capabilities, and functionality. On the right, a snapshot of our companion website for searching and exploring 3DSG publications, whose publication count highlights the field's exponential growth.}
\label{fig:timeline}
\end{center}
}]

\let\oldthefootnote\thefootnote
\renewcommand{\thefootnote}{}%
\footnotetext{
\textsuperscript{*}Co-led the project. \\
\texttt{dennis.rotondi@ki.uni-stuttgart.de},
\texttt{argenziano@diag.uniroma1.it}.}%
\let\thefootnote\oldthefootnote

\begin{abstract}
3D Scene Graphs (3DSGs) have emerged as a powerful representation for spatial AI by combining geometric grounding with semantic and relational abstractions of the environment. 
Their expressiveness has made them relevant to a broad range of problems in robotics and computer vision, including manipulation, navigation, task planning, scene understanding, and many others. 
However, the field remains fragmented: different communities adopt distinct formulations, construction pipelines, and evaluation protocols, making it difficult to compare methods, identify common assumptions, and assess remaining challenges for robust real-world deployment. 
This survey provides a unified and critical review of 3DSGs, with particular emphasis on open challenges and future directions.
We first formalize 3DSGs under a common definition and analyze the principal modeling choices that characterize existing formulations, including node and edge attributes, hierarchical structure, dynamic scene representations, and affordance-aware extensions.
We then review how 3DSGs are built from raw sensory observations, discussing the most common terminologies, conventions, and techniques. 
Finally, we examine downstream applications and evaluation strategies, from intrinsic graph quality to task-level performance.
To support the community, we also provide a dedicated website that organizes and extends the surveyed content, accessible at \website.
\end{abstract}

\section{Introduction}
\label{sec:intro}

Autonomous agents operating in unstructured real-world environments must be able to perceive, represent, and reason about complex three-dimensional scenes in ways that support both geometric accuracy (\eg to enable precise task execution) and semantic understanding (\eg to interpret and execute complex instructions). 
Classical spatial representations in robotics and computer vision~\cite{Cadena_2016,mascaro2025survey} provide precise geometric information but offer limited support for high-level reasoning, long-term memory, and interaction-centric tasks. Conversely, implicit language-based 
representations~\cite{lerf2023, lerftogo2023} capture rich semantic abstractions, yet lack explicit grounding in physical space. Bridging this gap between low-level geometry and high-level semantics remains a central challenge in spatial and embodied AI.

\tdsgslong (\tdsgsshort) \cite{armeni2019scenegraph, wald2020learning3d, rosinol2020dynamic} have emerged as a promising approach to address this challenge. By representing a scene as a graph, they provide a compact, interpretable, and queryable abstraction of 3D environments. 
This formulation naturally supports hierarchical organization, explicit relational reasoning, and multimodal grounding, while remaining anchored to an underlying geometric representation. The growing interest in \tdsgsshort is driven by several converging trends. 
First, advances in \ac{SLAM} and 3D reconstruction have enabled the creation of increasingly accurate, large-scale geometric models of real-world environments. 
Second, progress in object detection, instance segmentation, and multimodal \acp{FM} has significantly improved the extraction of semantic structure from visual data, including in open-vocabulary settings. 
Third, the rise of \acp{LLM} and \acp{VLM} has created a strong demand for structured world representations that can serve as interfaces between language, perception, and action.

However, despite the rapid progress, the field of \tdsgsshort remains fragmented. There is still no unified perspective that connects representation design, construction methodologies, downstream applications, and evaluation practices. 
This survey aims to address this gap by mapping the terminology used across communities, identifying how different terms often refer to the same underlying concepts in different use cases, and proposing a consistent conceptual framework.

We organize this review around four central questions related to \tdsgsshort:
\begin{itemize}
    \item \textbf{What is a \tdsgshort?}  
    We revisit its core formulation while clarifying the role of recent extensions such as hierarchical, dynamic, and interaction-aware representations.
    
    \item \textbf{How are \tdsgsshort built?}  
    We examine the main processing paradigms, input modalities, and key challenges that arise in real-world settings.
    
    \item \textbf{How are \tdsgsshort used and evaluated?}  
    We analyze the main downstream applications as well as the diverse %
    evaluation protocols adopted across communities.

    \item \textbf{What are the open challenges and future directions?}
    We identify gaps in current modeling choices, construction pipelines, applications, and benchmarking practices, and outline directions toward robust, general-purpose deployment.
\end{itemize}

As illustrated in~\Cref{fig:timeline}, we further contextualize these developments within the broader evolution of the field, emphasizing key milestones that have shaped the transition from early structural representations toward their use in planning and as a foundation for physical AI.
Finally, we release a novel web platform~\website~to facilitate the exploration and curation of \tdsgshort papers based on keywords derived from this survey.
Given the young age of the field, we aimed to be exhaustive and included all available relevant works in this review; the website, which is kept continuously updated, already covers more than 150 papers at the time of submission.

\section{Relation to Prior Surveys and Scope}
\label{sec:related}

\tdsgsshort have been adopted in a wide range of computer vision and robotics applications, where they are typically discussed as part of broader surveys on related technologies.

Scene Graphs were initially introduced as formal representations of images~\cite{krishna2017visual}, and the earliest comprehensive surveys~\cite{chang2021comprehensive, li2024scene} primarily focused on their role in 2D scene understanding, offering limited insights into the challenges posed by 3D environments such as partial observability, metric grounding, and scalability to large scenes.
Subsequently, \citet{bae2023survey3dscenegraphs} presented a concise review that shifted attention from 2D to \tdsgslong, highlighting their emerging potential as spatial representations, while covering only the earliest contributions in the field~\cite{armeni2019scenegraph, kim2019sparse3d, rosinol2021kimera, hughes2022hydra}.
A separate line of related surveys investigates scene representations~\cite{sh-ch17-openworld, garg2020semantics, mascaro2025survey} and \ac{SLAM}~\cite{rosen2021surveyslam, slam-handbook}, within which \tdsgslong are discussed as a structured extension of geometric mapping. These works emphasize the integration of object- and relation-level abstractions into traditional metric representations by introducing hierarchical structures that augment point- or surface-based maps, yet they approach \tdsgslong primarily from the perspective of spatial reconstruction, leaving their semantic and relational aspects underexplored.
\citet{ma2025llmsstep3dworld} survey \acp{LLM} for 3D perception and reasoning, highlighting \tdsgslong as a structured, easily parsable representation that supports high-level reasoning and language-based interaction.
Finally, \tdsgslong are increasingly discussed in the context of world models \cite{kong20253d4dworldmodeling}, which aim to learn persistent and predictive representations of 3D environments. 
While current world modeling research primarily relies on implicit representations, \tdsgslong provide explicit semantic and geometric grounding that can be used to condition such models.

Despite the growing applications of \tdsgsshort, no survey has yet provided a detailed analysis of its structure, applications, evaluation, and practical limitations. 

The most closely related effort is the concurrent work of \citet{catalano20253dsginrobotics}, which provides a taxonomy and broad overview of the state of the art. However, its primary aim is to define and categorize \tdsgslong, rather than to critically assess what is missing. 

Our survey differs in both intent and scope: we provide an in-depth analysis of \tdsgshort structures, their downstream applications, and, crucially, their evaluation protocols. 
Beyond categorizing existing methods, our goal is to highlight the main directions of ongoing community efforts, discuss the limitations of existing benchmarks and metrics, and outline the key open challenges that must be addressed to fully exploit the potential of \tdsgslong in real-world applications. 

While this manuscript focuses on the most representative papers in the area, we provide an interactive and comprehensive database at \website, where curated \tdsgshort publications can be searched using various keywords.

\section{What is a 3D Scene Graph?}
\label{sec:whatis3dsg}

\begin{figure*}
    \centering
    \includegraphics[width=\textwidth]{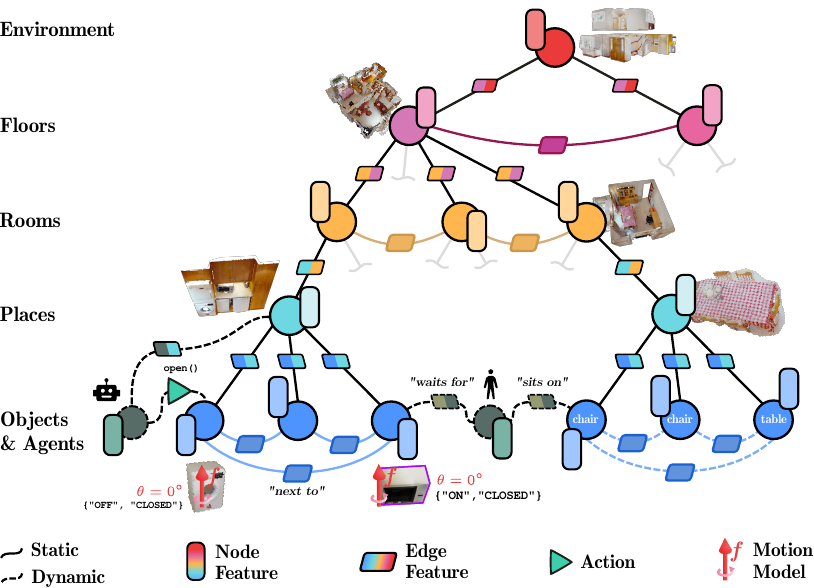}
   \caption{
\textbf{Example illustration of a hierarchical 3DSG for indoor environments. }
In this example, the scene is organized into five layers (environment, floors, rooms, places, and objects \& agents), encoding spatial containment via parent-child edges. 
Places decompose rooms into finer-grained functional regions (\eg the counter area of a kitchen or the area around a dining table). 
At the object layer, edges capture spatial and semantic relations between entities (\eg ``next to'', ``sits on'', ``waits for''), while node and edge features encode geometric, semantic, and textual attributes (\eg \{``OFF'', ``CLOSED''\}). 
Agent nodes (human or robot) operate at the object layer and interact with objects via actions, represented as edges with action attributes (\eg \texttt{open()}); an action's effect is realized as an update to the object's time-varying attributes (\eg \{``CLOSED''\} $\rightarrow$ \{``OPEN''\}).
Articulated objects are associated with a motion model that describes part-level motion (\eg joint angle $\theta$). 
Dashed lines denote time-varying components, whereas solid lines denote static structure.
}
    \label{fig:example3dsg}
\end{figure*}

\tdsgslong are a structured representation of metric and semantic information in 3D scenes.
Due to the generality of this concept, a variety of closely related \tdsgshort formulations have been proposed.
This section unifies them under a single definition and highlights how existing variants relate to it, with particular attention to the graph's structure and what the nodes and edges capture.
We define a \tdsglong as a tuple,
\begin{equation}
  \m{G} \;=\; \bigl(\,\m{V},\;\; V_g,\;\; \m{V}_{f},\;\; \m{E},\;\; \m{E}_{f},\;\; [\,\m{L}\,]\,\bigr),
  \label{eq:dsg}
\end{equation}
where $\m{V}$ and $\m{E}$ are the node and edge sets, $\m{V}_g$ \emph{grounds} each node in the 3D map, $\m{V}_{f}$ and $\m{E}_{f}$ assign features (or attributes) to nodes and edges, and $\m{L}$ is an optional labeling function organizing nodes into hierarchical layers. 
We describe each component in turn.

The grounding $\m{V}_g : \m{V} \to \m{M}$, with $\m{M}\subseteq \mathbb{R}^3$, assigns to each node $v$ the subset of metric map points it occupies.
Each node $v_i \in \m{V}$ thus represents an entity or concept \emph{grounded in 3D space}, often an abstraction of low-level scene primitives and geometry (\eg objects or object parts) or spatial regions (\eg rooms or floors). 
Note that the grounding does not constrain the scene representation to be a point cloud. Rather, point clouds serve as a common geometric denominator to which other reconstruction modalities, such as meshes~\cite{armeni2019scenegraph, rosinol2021kimera} and 3D Gaussian splatting models~\cite{ge2025dynamicgsgd}, can be reduced.

Nodes also carry attributes through the feature map $\m{V}_{f} : \m{V} \to \m{V}_{\m{A}}$.
These are broadly categorized into \emph{geometric attributes}, such as the centroid, enclosing bounding box, or 3D shape, and \emph{semantic attributes}, such as class labels, material properties, color~\cite{armeni20163d}, multimodal embeddings~\cite{chen2024clip}, or textual descriptions~\cite{gorlo2025describe}.

Each edge $e_{ij} \in \m{E}$ represents a directed relationship $(v_i \xrightarrow{} v_j)$ between two nodes $v_i, v_j \in \m{V}$, though this can be relaxed to undirected edges for symmetric relations.
Multiple edges between the same pair of nodes are permitted, yielding a \emph{multigraph}~\cite{agia2022taskography}.

Like nodes, edges carry attributes through an edge feature map  $\m{E}_{f} : \m{E} \to \m{E}_{\m{A}}$. 
These attributes capture the nature of the relationship between connected nodes, which may be spatial (\eg ``next to'', ``contained in''), structural (\eg ``is part of''), semantic (\eg ``smaller than''), or functional (\eg a switch ``controls'' a light, or an
action's effect). 
When the focus is specifically on semantic relationships, the graph is commonly referred to as a 3D Semantic Scene Graph (3DSSG)~\cite{wald2020learning3d, wang2023vlsat, saxena2024grapheqa, ozsoy2024holistic, hou2025fross}.

The labeling function $\m{L} : \m{V} \to \mathbb{N}$ organizes nodes into hierarchical \emph{layers}, assigning each node to a level (\eg objects within rooms); a flat \tdsgshort simply omits it.
\Cref{sec:whatis3dsg:modeling} develops the layered structure and how the layers are chosen for indoor and outdoor scenes.

Finally, any part of~\Cref{eq:dsg} may be indexed by time $t$.
Attributes may therefore be static (\eg a class label or fixed geometry) or time-varying (\eg an object's state, a door's opening angle $\theta$, or the interval over which a relation holds). 
In this context, a temporal grounding function $\m{V}_g^{[t]}$ can capture the motion or deformation of a node over time.
A \tdsgshort with any time-indexed component is a \emph{4D Scene Graph}; we treat dynamic and 4D representations, and the beliefs they require, in \Cref{sec:modelingdynamics}.

Existing \tdsgshort variants can be understood as instances of this framework, differing primarily in their choices of nodes, edges, features, layers, and temporal modeling. \Cref{fig:example3dsg} illustrates these components in an example indoor 3DSG.

\subsection{Modeling Hierarchies}
\label{sec:whatis3dsg:modeling}
A hierarchical \tdsgshort instantiates the optional labeling function $\m{L} : \m{V} \to \mathbb{N}$ introduced above, which assigns each node $v_i \in \m{V}$ to a layer $l \in \mathbb{N}$.
This induces a partition of the node set
\(
  \m{V} = \bigcup_{l=1}^{L} \m{V}_k,
\)
where $\m{V}_l = \{\, v \in \m{V} \mid \m{L}(v) = l \,\}$.
Without loss of generality, we define layer $l = L$ as the layer containing the root of the graph, and layer $l = 1$ as the layer encoding the lowest-level, often metric information (\eg meshes or \pcds).
The edge set \(\m{E}\) divides into
\(
  \m{E} = \m{E}_{\mathrm{H}} \;\cup\; \m{E}_{\mathrm{R}},
\)
where
\(
\m{E}_{\mathrm{H}} \subseteq \bigcup_{l=1}^{L-1} (\m{V}_l \times \m{V}_{l+1})\) are the hierarchical edges connecting adjacent layers, and edges in $\m{E}_{\mathrm{R}} \subseteq \m{V} \times \m{V}$ relate any pair of nodes, but in practice predominantly relate nodes within the same layer, and are therefore commonly referred to as intra-layer relationships.
Beyond yielding expressive and interpretable representations, arranging information hierarchically offers memory and computational benefits: \eg \citet{hughes2024foundations} show that, under certain assumptions, the \tdsgsshort underlying indoor spaces have low treewidth, enabling fast inference.

The hierarchy's layers are often defined based on the scene in which the agent will operate. 
A popular choice for indoor environments consists of a building layer as the root node that then descends into rooms, places, and objects~\cite{armeni2019scenegraph, ravichandran2022hierarchical, hughes2022hydra, maggio2024clio, xu2025tb}. 
In multi-floor buildings, there can also be a floor layer between the root and the rooms~\cite{agia2022taskography, werby2024hierarchical, werby2025keysg, linok2025indoorgrounding}. 
In Situational Graphs~\cite{bavle2022situational, bavle2023sgraphs, bavle2025sgraphs2, millanromera2025metricsemanticfactorgraph} the object layer is replaced by a layer of geometric primitives, such as wall and corridor planes, due to the lack of low-level semantics.

In contrast, outdoor hierarchies lack the rigid structural boundaries found indoors, necessitating a shift toward more topological or functional decomposition. 
For urban environments, the root node typically represents a city district or neighborhood, with subsequent layers consisting of roads and intersections, lane graphs, and object instances~\cite{greve2024collaborative, steinke2025curbosg, deng2024opengraph}. 
An alternative, functional way to define the hierarchy consists of splitting the environment into distinct terrain regions (\eg ``sidewalk'', ``grass'', ``asphalt'')~\cite{samuelson2025terra, samuelson2025terrainawaretaskdriven, stathoulopoulos2025havewe} or other high-level concepts such as ``orchards'', ``verges'', or ``footings''~\cite{strader2024spatial, xie2024embodied, viswanathan2024xflieleveragingactionablehierarchical, Nyffeler_2025_ICCV, yang2025h3dsgplant}.

\boldparagraph{Open Challenges: Defining and Inferring Abstract Layers} While progress has been made in representing hierarchical partitions for indoor and outdoor environments per se, only \citet{strader2024spatial} attempt to address both domains simultaneously within a general framework. Determining the ideal general spatial partition remains difficult: it is often challenging to obtain meaningful high-level regions in general environments, as the lack of structural boundaries in large open areas forces a reliance on subjective or functional criteria, suggesting that a more adaptive approach is needed. 
At a more fundamental level, the definition of layers is often hard-coded and inflexible. 
Although recent works, such as \cite{chang2025ashita}, have begun to infer layer definitions automatically in a task-aware manner, a more rigorous and flexible formulation of abstraction layers is still lacking.

\subsection{Modeling Dynamics}
\label{sec:modelingdynamics}

In many real-world scenarios, autonomous agents must account for a variety of dynamic environmental effects, ranging from moving vehicles and gradual scene changes in outdoor settings to humans interacting with objects at home.
Capturing both \emph{short-term} dynamics (\eg a person walking through the view of the robot) and \emph{long-term} changes (\eg items that get displaced when they are outside the view of the robot, see \cite{sh-ch15-dyndef} for definitions and discussion) is essential for human-centric and long-term autonomy.

Our definition accommodates this with no new machinery (its components may simply vary with time), yet representing such variation well poses several significant challenges. 
First, one needs a \emph{representation} of dynamic entities in the scene, often referred to as \emph{Dynamic} \tdsgslong (DSGs\footnote{The acronyms DSG and \tdsgshort (sometimes also only DSG) are used in the literature to refer to \emph{dynamic}~SGs~\cite{rosinol2020dynamic} and 3D~SGs, respectively. The similarity is unfortunate, and funnily enough, another source of confusion is the acronym 3DGS (3D Gaussian Splatting).}) \cite{rosinol2021kimera, greve2024collaborative, ge2025dynamicgsgd, steinke2025curbosg}.
This can be implemented in a variety of ways, with complementary limitations and benefits. 
One approach is to instantiate a separate \tdsgshort for each timestamp $t$, producing a sequence of graphs $\m{G}^{[1]}, \dots, \m{G}^{[T]}$, each a full instance of~\Cref{eq:dsg}.
This representation provides a simple interface for learning~\cite{wang2023trajectory}, but duplicates large amounts of \emph{static} information.
An alternative is to create distinct nodes for different observations or temporal instances of the same entity~\cite{schmid2024khronos}: this preserves the observation history and ensures temporal consistency, but requires an additional mechanism to reason about which nodes correspond to the same physical entity.
The most common approach in practice is to maintain nodes and edges with time-varying attributes~\cite{rosinol2021kimera,greve2024collaborative, ge2025dynamicgsgd, steinke2025curbosg}. 
This results in compact representations, but requires accurate data-association to avoid corrupting the attributes through changes.
This is thus particularly suitable for short-term dynamics, where agent-specific attributes can capture the unique kinematics of dynamic entities.
For instance, human nodes~\cite{rosinol2020dynamic, rosinol2021kimera} can store specialized representations such as skeletal joints and pelvis locations, illustrating how node attributes are extended to capture agent-specific kinematics.

Second, dynamic scenes can quickly make their representations inaccurate or outdated.
On the one hand, \emph{long-term} dynamics require mechanisms to integrate the observed changes into the \tdsgshort representation~\cite{yan2025dynamicopenvocabulary, nguyen2025efficientattribute}. 
Beyond keeping the representation up to date, some approaches reconstruct the scene's \emph{history}, recording how it evolved over time~\cite{schmid2024khronos, gorlo2025describe}. 
On the other hand, an ideal 4D representation should also look forward, anticipating the scene's plausible future states.
For instance, some approaches encode the probability of different long-term object configurations~\cite{looper2023vsg}, or to predict future actions~\cite{patel2022proactive} and trajectories~\cite{gorlo2024trajectory} of agents, based on their spatial and semantic context in a \tdsgshort.

Finally, since \tdsgsshort excel at capturing not only entities but their relations as well as higher-level spatial abstractions, a complete generalization of \tdsgsshort to 4DSGs should also capture spatio-temporal relationships as well as temporal abstractions.
While recent initial work has investigated, \eg maps-of-dynamics (MoD)-based approaches to abstract trajectories in 4DSGs~\cite{catalano2025aion}, this area remains largely underexplored.

\boldparagraph{Open Challenges: Scaling, Uncertainty, and World Modeling}
Several fundamental challenges remain: current representations often grow with the amount of observed data and have primarily been demonstrated over limited durations. 
Developing compact representations and compression, summarization, or marginalization strategies that enable operation over months or years is an important open problem. 
Then, most approaches maintain a single estimate of the world state, despite the inherent uncertainty arising from partial observability, ambiguous data association, and future scene evolution. How to represent and update beliefs over past, present, and future states in 4DSGs has received little attention. Finally, most existing work focuses on tracking entities through time, whereas the future evolution is comparatively understudied. Designing predictive models that capture how entities and relationships emerge, disappear, and evolve as objects and agents act and interact is an important direction for future work.

\subsection{Modeling Functionality and Actionability}
\label{sec:modelingfunctionality}
The content of the nodes and edges can be specialized to capture how agents interact and act within their environment.
For example, \emph{functional} \tdsgslong \cite{rotondi2025fungraph, zhang2025functional3dsg, engelbracht2024spotlight, gu2025artisgfunctional3dscene, buechner2026momasg, Fu_2026_funfact, feng2026tfuns3dtask} model the internal functional structure of objects by representing interactive sub-parts (such as knobs and handles) as separate nodes, optionally enriched with kinematic attributes \cite{gu2025artisgfunctional3dscene, buechner2026momasg} that store the current object state (\eg the opening angle).
Edges encode both intra-object composition (\eg a knob is part of an oven) and functional relationships (\eg a knob adjusts the oven's temperature). 
Notably, these functional edges go beyond Gibsonian affordances (\eg a button can be pressed), which earlier works represent as node attributes~\cite{wald2020learning3d}, and instead capture telic affordances~\cite{delitzas2024scenefun3d}, representing the behavior of an object’s function (\eg pressing the button turns on the light).

Focusing on physical interaction, \citet{jiang2024roboexp} propose \emph{Action-Conditioned} \tdsgsshort inspired by \ac{POMDP} notation, which introduce \emph{action} nodes representing the conditional steps required to interact with an object. 
Edges connecting object and action nodes encode the logical sequence of action effects. 
For example, opening a cabinet door to reveal a roll of tape is represented by an edge from the ``cabinet door'' node to the ``open'' action node, followed by an edge from ``open'' to the ``tape'' node. 
Building on this formulation, \citet{wang2025curiousbot} extend the approach to more complex interactions, such as flipping a box to uncover a hidden toy.
Within our framework, actions may be represented either as nodes, by grounding them to the spatial region in which they occur, or as edges with action attributes (\eg ``open'').

While these approaches focus on \emph{how} actions are executed, it is possible to instantiate special nodes that support \emph{where} to act and \emph{what} to reason about.
In particular, \tdsgsshort can have nodes corresponding to traversable or free space, with traversability between them encoded as edges \cite{kim2019sparse3d, hughes2022hydra, rosinol2020dynamic}. 
Similarly, ``blind nodes'' \cite{saucedo2025estimatingcommonsense, saucedo2024beliefsg} represent beliefs over entirely unobserved portions of the scene, such as objects likely to co-occur with those already detected in a given room. 
Nodes can also capture entities that have been partially perceived but whose properties remain incomplete, supporting tasks like exploration where uncertain information must still guide planning \cite{viswanathan2024xflieleveragingactionablehierarchical, viswanathan2025spade}.

\boldparagraph{Open Challenges: Task Awareness, Physics, and Causality}
While this family of graphs is highly versatile, representing all object sub-parts, actions, beliefs, and functional relationships can result in large, computationally expensive graphs that are impractical for resource-constrained embodied platforms. 
A key open direction is the development of lightweight parametric nodes with mechanisms that dynamically expand or compress functional detail based on task and context. 
Building on such compact representations, node attributes could be extended to include dynamic parameters such as mass and material properties, enabling more accurate adherence to payload limits and force requirements during manipulation. 
At the same time, actions continuously modify the scene and its functional structure: opening a cabinet reveals previously inaccessible objects, while moving or removing an object can change the affordances and physical stability of surrounding objects. Maintaining a causally consistent graph that captures and propagates these changes over long interaction horizons remains an open challenge.

\section{How are 3D Scene Graphs built?}
\label{sec:constructing}

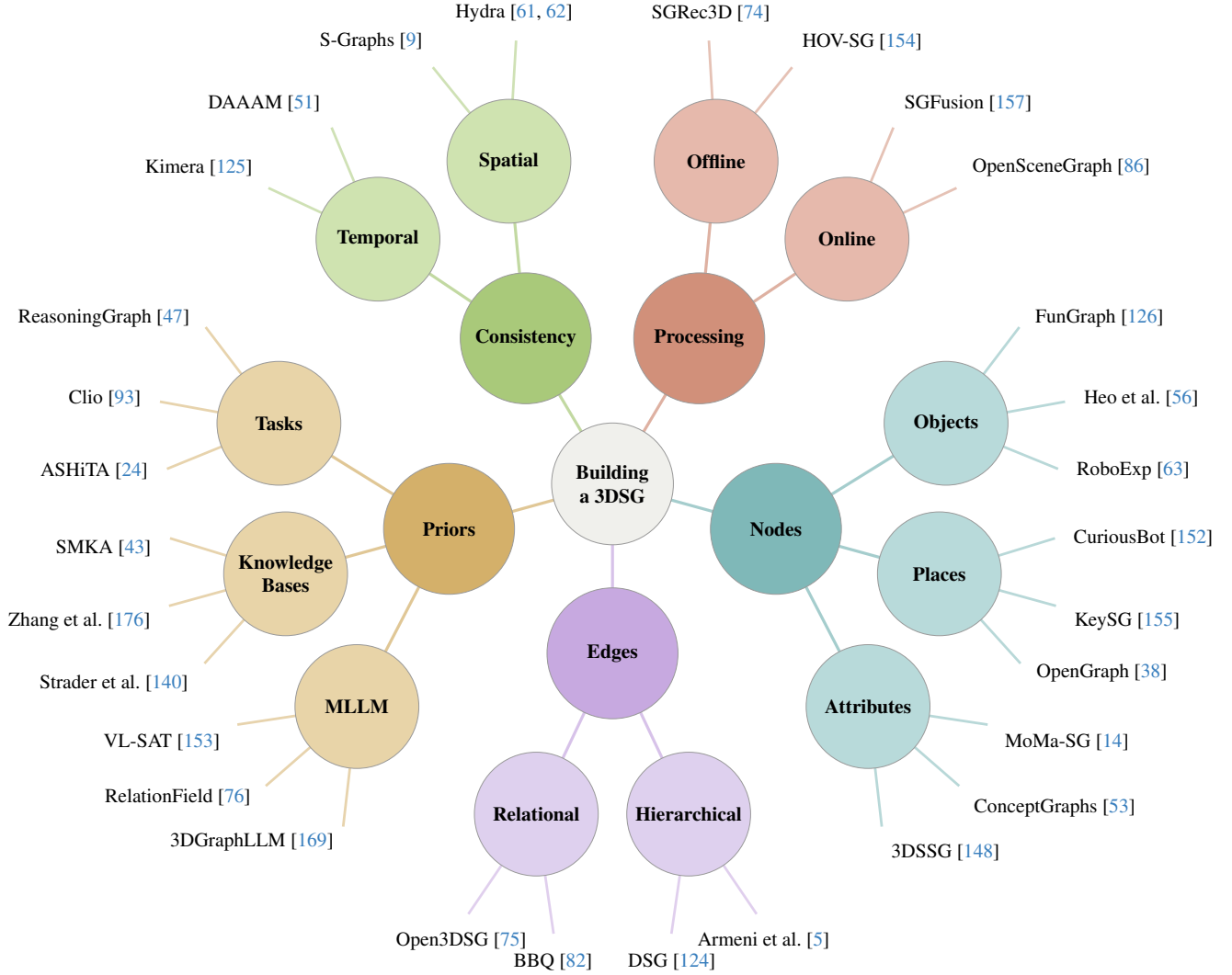
\begin{figure*}
    \centering
    \resizebox{\textwidth}{!}{%
\begin{tikzpicture}[
    every node/.style={inner sep=0pt, outer sep=0pt},
]

\definecolor{cF0F0EC}{RGB}{240,240,236}
\definecolor{cD1907A}{RGB}{209,144,122}
\definecolor{cE6B9AC}{RGB}{230,185,172}
\definecolor{cF6EDEA}{RGB}{246,237,234}
\definecolor{c7FB8B8}{RGB}{127,184,184}
\definecolor{cB8DADA}{RGB}{184,218,218}
\definecolor{cE9F4F4}{RGB}{233,244,244}
\definecolor{cC7A9E0}{RGB}{199,169,224}
\definecolor{cDED0EE}{RGB}{222,208,238}
\definecolor{cF4EEFA}{RGB}{244,238,250}
\definecolor{cD4B06A}{RGB}{212,176,106}
\definecolor{cE8D4A8}{RGB}{232,212,168}
\definecolor{cF7F1E5}{RGB}{247,241,229}
\definecolor{cA9C97A}{RGB}{169,201,122}
\definecolor{cCFE3B0}{RGB}{207,227,176}
\definecolor{cEEF5E6}{RGB}{238,245,230}

\definecolor{edgeProcessing}{RGB}{209,144,122}
\definecolor{edgeNodes}{RGB}{127,184,184}
\definecolor{edgeEdges}{RGB}{199,169,224}
\definecolor{edgePriors}{RGB}{212,176,106}
\definecolor{edgeConsistency}{RGB}{169,201,122}

\draw[edgeProcessing, line width=1.50pt, opacity=0.70] (0.0000,0.0000) -- (1.5381,2.5826);
\draw[edgeProcessing, line width=1.50pt, opacity=0.70] (1.5381,2.5826) -- (1.8367,5.7245);
\draw[edgeProcessing, line width=1.30pt, opacity=0.55, shorten >=3pt] (1.8367,5.7245) -- (3.2524,7.4734);
\draw[edgeProcessing, line width=1.30pt, opacity=0.55, shorten >=3pt] (1.8367,5.7245) -- (1.7030,7.9706);
\draw[edgeProcessing, line width=1.50pt, opacity=0.70] (1.5381,2.5826) -- (4.1584,4.3420);
\draw[edgeProcessing, line width=1.30pt, opacity=0.55, shorten >=3pt] (4.1584,4.3420) -- (6.1969,5.2943);
\draw[edgeProcessing, line width=1.30pt, opacity=0.55, shorten >=3pt] (4.1584,4.3420) -- (5.0216,6.4197);

\draw[edgeNodes, line width=1.50pt, opacity=0.70] (0.0000,0.0000) -- (2.8982,-0.7976);
\draw[edgeNodes, line width=1.50pt, opacity=0.70] (2.8982,-0.7976) -- (5.9152,1.0751);
\draw[edgeNodes, line width=1.30pt, opacity=0.55, shorten >=3pt] (5.9152,1.0751) -- (7.9987,0.2259);
\draw[edgeNodes, line width=1.30pt, opacity=0.55, shorten >=3pt] (5.9152,1.0751) -- (8.1288,1.4774);
\draw[edgeNodes, line width=1.30pt, opacity=0.55, shorten >=3pt] (5.9152,1.0751) -- (7.2785,2.8651);
\draw[edgeNodes, line width=1.50pt, opacity=0.70] (2.8982,-0.7976) -- (5.7965,-1.5953);
\draw[edgeNodes, line width=1.30pt, opacity=0.55, shorten >=3pt] (5.7965,-1.5953) -- (7.3058,-3.2639);
\draw[edgeNodes, line width=1.30pt, opacity=0.55, shorten >=3pt] (5.7965,-1.5953) -- (7.9659,-2.1924);
\draw[edgeNodes, line width=1.30pt, opacity=0.55, shorten >=3pt] (5.7965,-1.5953) -- (7.9472,-0.9340);
\draw[edgeNodes, line width=1.50pt, opacity=0.70] (2.8982,-0.7976) -- (4.5320,-3.9503);
\draw[edgeNodes, line width=1.30pt, opacity=0.55, shorten >=3pt] (4.5320,-3.9503) -- (4.7876,-6.1857);
\draw[edgeNodes, line width=1.30pt, opacity=0.55, shorten >=3pt] (4.5320,-3.9503) -- (6.2280,-5.4288);
\draw[edgeNodes, line width=1.30pt, opacity=0.55, shorten >=3pt] (4.5320,-3.9503) -- (6.7567,-4.2867);

\draw[edgeEdges, line width=1.50pt, opacity=0.70] (0.0000,0.0000) -- (0.0000,-3.0060);
\draw[edgeEdges, line width=1.50pt, opacity=0.70] (0.0000,-3.0060) -- (1.3511,-5.8583);
\draw[edgeEdges, line width=1.30pt, opacity=0.55, shorten >=3pt] (1.3511,-5.8583) -- (1.0296,-8.0852);
\draw[edgeEdges, line width=1.30pt, opacity=0.55, shorten >=3pt] (1.3511,-5.8583) -- (2.6152,-7.7195);
\draw[edgeEdges, line width=1.50pt, opacity=0.70] (0.0000,-3.0060) -- (-1.3511,-5.8583);
\draw[edgeEdges, line width=1.30pt, opacity=0.55, shorten >=3pt] (-1.3511,-5.8583) -- (-2.6152,-7.7195);
\draw[edgeEdges, line width=1.30pt, opacity=0.55, shorten >=3pt] (-1.3511,-5.8583) -- (-1.0296,-8.0852);

\draw[edgePriors, line width=1.50pt, opacity=0.70] (0.0000,0.0000) -- (-2.8982,-0.7976);
\draw[edgePriors, line width=1.50pt, opacity=0.70] (-2.8982,-0.7976) -- (-4.5320,-3.9503);
\draw[edgePriors, line width=1.30pt, opacity=0.55, shorten >=3pt] (-4.5320,-3.9503) -- (-6.7567,-4.2867);
\draw[edgePriors, line width=1.30pt, opacity=0.55, shorten >=3pt] (-4.5320,-3.9503) -- (-6.2280,-5.4288);
\draw[edgePriors, line width=1.30pt, opacity=0.55, shorten >=3pt] (-4.5320,-3.9503) -- (-4.7876,-6.1857);
\draw[edgePriors, line width=1.50pt, opacity=0.70] (-2.8982,-0.7976) -- (-5.7965,-1.5953);
\draw[edgePriors, line width=1.30pt, opacity=0.55, shorten >=3pt] (-5.7965,-1.5953) -- (-7.9472,-0.9340);
\draw[edgePriors, line width=1.30pt, opacity=0.55, shorten >=3pt] (-5.7965,-1.5953) -- (-7.9659,-2.1924);
\draw[edgePriors, line width=1.30pt, opacity=0.55, shorten >=3pt] (-5.7965,-1.5953) -- (-7.3058,-3.2639);
\draw[edgePriors, line width=1.50pt, opacity=0.70] (-2.8982,-0.7976) -- (-5.9152,1.0751);
\draw[edgePriors, line width=1.30pt, opacity=0.55, shorten >=3pt] (-5.9152,1.0751) -- (-7.2785,2.8651);
\draw[edgePriors, line width=1.30pt, opacity=0.55, shorten >=3pt] (-5.9152,1.0751) -- (-8.1288,1.4774);
\draw[edgePriors, line width=1.30pt, opacity=0.55, shorten >=3pt] (-5.9152,1.0751) -- (-7.9987,0.2259);

\draw[edgeConsistency, line width=1.50pt, opacity=0.70] (0.0000,0.0000) -- (-1.5381,2.5826);
\draw[edgeConsistency, line width=1.50pt, opacity=0.70] (-1.5381,2.5826) -- (-4.1584,4.3420);
\draw[edgeConsistency, line width=1.30pt, opacity=0.55, shorten >=3pt] (-4.1584,4.3420) -- (-5.0216,6.4197);
\draw[edgeConsistency, line width=1.30pt, opacity=0.55, shorten >=3pt] (-4.1584,4.3420) -- (-6.1969,5.2943);
\draw[edgeConsistency, line width=1.50pt, opacity=0.70] (-1.5381,2.5826) -- (-1.8367,5.7245);
\draw[edgeConsistency, line width=1.30pt, opacity=0.55, shorten >=3pt] (-1.8367,5.7245) -- (-1.7030,7.9706);
\draw[edgeConsistency, line width=1.30pt, opacity=0.55, shorten >=3pt] (-1.8367,5.7245) -- (-3.2524,7.4734);

\node[text=black, anchor=south west, inner sep=1pt, font=\fontsize{10}{12}\selectfont] at (3.3322,7.6568) {HOV-SG~\cite{werby2024hierarchical}};
\node[text=black, anchor=south, inner sep=1pt, font=\fontsize{10}{12}\selectfont] at (1.7448,8.1662) {SGRec3D~\cite{koch2024sgrec3d}};
\node[text=black, anchor=south west, inner sep=1pt, font=\fontsize{10}{12}\selectfont] at (6.3489,5.4243) {OpenSceneGraph~\cite{loo2024openscenegraphsopen}};
\node[text=black, anchor=south west, inner sep=1pt, font=\fontsize{10}{12}\selectfont] at (5.1449,6.5772) {SGFusion~\cite{wu2021scenegraphfusion}};
\node[text=black, anchor=west, inner sep=1pt, font=\fontsize{10}{12}\selectfont] at (8.1986,0.2315) 
{RoboExp~\cite{jiang2024roboexp}};
\node[text=black, anchor=west, inner sep=1pt, font=\fontsize{10}{12}\selectfont] at (8.3256,1.5132) {Heo et al.~\cite{heo2025object}};
\node[text=black, anchor=west, inner sep=1pt, font=\fontsize{10}{12}\selectfont] at (7.4646,2.9383) {FunGraph~\cite{rotondi2025fungraph}};
\node[text=black, anchor=north west, inner sep=1pt, font=\fontsize{10}{12}\selectfont] at (8.1587,-2.2455) {KeySG~\cite{werby2025keysg}};
\node[text=black, anchor=west, inner sep=1pt, font=\fontsize{10}{12}\selectfont] at (7.4884,-3.3455) {OpenGraph~\cite{deng2024opengraph}};
\node[text=black, anchor=west, inner sep=1pt, font=\fontsize{10}{12}\selectfont] at (8.1458,-0.9574) {CuriousBot~\cite{wang2025curiousbot}};
\node[text=black, anchor=north west, inner sep=1pt, font=\fontsize{10}{12}\selectfont] at (4.9101,-6.3439) {3DSSG~\cite{wald2020learning3d}};
\node[text=black, anchor=north west, inner sep=1pt, font=\fontsize{10}{12}\selectfont] at (6.3788,-5.5602) {ConceptGraphs~\cite{gu2024conceptgraphs}};
\node[text=black, anchor=north west, inner sep=1pt, font=\fontsize{10}{12}\selectfont] at (6.9255,-4.3938) {MoMa-SG~\cite{buechner2026momasg}};
\node[text=black, anchor=north, inner sep=1pt, font=\fontsize{10}{12}\selectfont] at (2.6794,-7.9089) {Armeni et al.~\cite{armeni2019scenegraph}};
\node[text=black, anchor=north, inner sep=1pt, font=\fontsize{10}{12}\selectfont] at (1.0549,-8.2836) {DSG~\cite{rosinol2020dynamic}};
\node[text=black, anchor=north, inner sep=1pt, font=\fontsize{10}{12}\selectfont] at (-1.0549,-8.2836) {BBQ~\cite{linok2025beyondbarequeries}};
\node[text=black, anchor=north, inner sep=1pt, font=\fontsize{10}{12}\selectfont] at (-2.6794,-7.9089)  {Open3DSG~\cite{koch2024open3dsg}};

\node[text=black, anchor=north east, inner sep=1pt, font=\fontsize{10}{12}\selectfont] at (-8.1458,-0.9574) {SMKA~\cite{feng2023spatial}};

\node[text=black, anchor=north east, inner sep=1pt, font=\fontsize{10}{12}\selectfont] at (-7.4884,-3.3455) {Strader et al.~\cite{strader2024spatial}};

\node[text=black, anchor=north east, inner sep=1pt, font=\fontsize{10}{12}\selectfont] at (-8.1587,-2.2455)(-8.1458,-0.9574) {Zhang et al.~\cite{zhang2021knowledge}};

\node[text=black, anchor=east, inner sep=1pt, font=\fontsize{10}{12}\selectfont] at (-4.9101,-6.3439) {3DGraphLLM~\cite{zemskova20253dgraphllm}};

\node[text=black, anchor=east, inner sep=1pt, font=\fontsize{10}{12}\selectfont] at (-6.3788,-5.5602) {RelationField~\cite{koch2025relationfield}};

\node[text=black, anchor=north east, inner sep=1pt, font=\fontsize{10}{12}\selectfont] at (-6.9255,-4.3938) {VL-SAT~\cite{wang2023vlsat}};

\node[text=black, anchor=east, inner sep=1pt, font=\fontsize{10}{12}\selectfont] at (-7.4646,2.9383) {ReasoningGraph~\cite{puigjaner2026reasoninggraph}}; 
\node[text=black, anchor=east, inner sep=1pt, font=\fontsize{10}{12}\selectfont] at (-8.3256,1.5132) {Clio~\cite{maggio2024clio}};
\node[text=black, anchor=east, inner sep=1pt, font=\fontsize{10}{12}\selectfont] at (-8.1986,0.2315) {ASHiTA~\cite{chang2025ashita}};
\node[text=black, anchor=south east, inner sep=1pt, font=\fontsize{10}{12}\selectfont] at (-5.1449,6.5772) {DAAAM~\cite{gorlo2025describe}};
\node[text=black, anchor=south east, inner sep=1pt, font=\fontsize{10}{12}\selectfont] at (-6.3489,5.4243) {Kimera~\cite{rosinol2021kimera}};
\node[text=black, anchor=south, inner sep=1pt, font=\fontsize{10}{12}\selectfont] at (-1.7448,8.1662) {Hydra~\cite{hughes2022hydra, hughes2024foundations}};
\node[text=black, anchor=south east, inner sep=1pt, font=\fontsize{10}{12}\selectfont] at (-3.3322,7.6568) {S-Graphs~\cite{bavle2022situational}};
\filldraw[fill=cE6B9AC, fill opacity=1.00, draw=black!40, line width=0.3pt, draw opacity=1.00] (1.8367,5.7245) circle (1.0980cm);
\node[text=black, font=\fontsize{10}{12}\selectfont\bfseries ] at (1.8367,5.7245) {Offline};
\filldraw[fill=cE6B9AC, fill opacity=1.00, draw=black!40, line width=0.3pt, draw opacity=1.00] (4.1584,4.3420) circle (1.0980cm);
\node[text=black, font=\fontsize{10}{12}\selectfont\bfseries ] at (4.1584,4.3420) {Online};
\filldraw[fill=cB8DADA, fill opacity=1.00, draw=black!40, line width=0.3pt, draw opacity=1.00] (5.9152,1.0751) circle (1.0980cm);
\node[text=black, font=\fontsize{10}{12}\selectfont\bfseries ] at (5.9152,1.0751) {Objects};
\filldraw[fill=cB8DADA, fill opacity=1.00, draw=black!40, line width=0.3pt, draw opacity=1.00] (5.7965,-1.5953) circle (1.0980cm);
\node[text=black, align=center, font=\fontsize{10}{12}\selectfont\bfseries ] at (5.7965,-1.5953) {Places};
\filldraw[fill=cB8DADA, fill opacity=1.00, draw=black!40, line width=0.3pt, draw opacity=1.00] (4.5320,-3.9503) circle (1.0980cm);
\node[text=black, font=\fontsize{10}{12}\selectfont\bfseries ] at (4.5320,-3.9503) {Attributes};
\filldraw[fill=cDED0EE, fill opacity=1.00, draw=black!40, line width=0.3pt, draw opacity=1.00] (1.3511,-5.8583) circle (1.0980cm);
\node[text=black, font=\fontsize{10}{12}\selectfont\bfseries ] at (1.3511,-5.8583) {Hierarchical};
\filldraw[fill=cDED0EE, fill opacity=1.00, draw=black!40, line width=0.3pt, draw opacity=1.00] (-1.3511,-5.8583) circle (1.0980cm);
\node[text=black, font=\fontsize{10}{12}\selectfont\bfseries ] at (-1.3511,-5.8583) {Relational};

\filldraw[fill=cE8D4A8, fill opacity=1.00, draw=black!40, line width=0.3pt, draw opacity=1.00] (-5.7965,-1.5953) circle (1.0980cm);
\node[text=black, align=center, font=\fontsize{10}{12}\selectfont\bfseries ] at (-5.7965,-1.5953) {Knowledge\\Bases};

\filldraw[fill=cE8D4A8, fill opacity=1.00, draw=black!40, line width=0.3pt, draw opacity=1.00] (-4.5320,-3.9503) circle (1.0980cm);
\node[text=black, font=\fontsize{10}{12}\selectfont\bfseries ] at (-4.5320,-3.9503) {MLLM};
\filldraw[fill=cE8D4A8, fill opacity=1.00, draw=black!40, line width=0.3pt, draw opacity=1.00] (-5.9152,1.0751) circle (1.0980cm);
\node[text=black, font=\fontsize{10}{12}\selectfont\bfseries ] at (-5.9152,1.0751) {Tasks};
\filldraw[fill=cCFE3B0, fill opacity=1.00, draw=black!40, line width=0.3pt, draw opacity=1.00] (-4.1584,4.3420) circle (1.0980cm);
\node[text=black, font=\fontsize{10}{12}\selectfont\bfseries ] at (-4.1584,4.3420) {Temporal};
\filldraw[fill=cCFE3B0, fill opacity=1.00, draw=black!40, line width=0.3pt, draw opacity=1.00] (-1.8367,5.7245) circle (1.0980cm);
\node[text=black, font=\fontsize{10}{12}\selectfont\bfseries ] at (-1.8367,5.7245) {Spatial};
\filldraw[fill=cD1907A, fill opacity=1.00, draw=black!40, line width=0.3pt, draw opacity=1.00] (1.5381,2.5826) circle (1.1610cm);
\node[text=black, font=\fontsize{10}{12}\selectfont\bfseries ] at (1.5381,2.5826) {Processing};
\filldraw[fill=c7FB8B8, fill opacity=1.00, draw=black!40, line width=0.3pt, draw opacity=1.00] (2.8982,-0.7976) circle (1.1610cm);
\node[text=black, font=\fontsize{10}{12}\selectfont\bfseries ] at (2.8982,-0.7976) {Nodes};
\filldraw[fill=cC7A9E0, fill opacity=1.00, draw=black!40, line width=0.3pt, draw opacity=1.00] (0.0000,-3.0060) circle (1.1610cm);
\node[text=black, font=\fontsize{10}{12}\selectfont\bfseries ] at (0.0000,-3.0060) {Edges};
\filldraw[fill=cD4B06A, fill opacity=1.00, draw=black!40, line width=0.3pt, draw opacity=1.00] (-2.8982,-0.7976) circle (1.1610cm);
\node[text=black, font=\fontsize{10}{12}\selectfont\bfseries ] at (-2.8982,-0.7976) {Priors};
\filldraw[fill=cA9C97A, fill opacity=1.00, draw=black!40, line width=0.3pt, draw opacity=1.00] (-1.5381,2.5826) circle (1.1610cm);
\node[text=black, font=\fontsize{10}{12}\selectfont\bfseries ] at (-1.5381,2.5826) {Consistency};
\filldraw[fill=cF0F0EC, fill opacity=1.00, draw=black!40, line width=0.3pt, draw opacity=1.00] (0.0000,0.0000) circle (1.0800cm);
\node[text=black, align=center, font=\fontsize{10}{12}\selectfont\bfseries ] at (0.0000,0.0000) {Building\\a 3DSG};

\end{tikzpicture}

    }
    \caption{
        \textbf{Conceptual map of 3DSG construction.} 
        The problem is organized along five dimensions that capture key design decisions: how observations are processed (processing), how entities are defined and represented (nodes), what relationships are established (edges), what auxiliary knowledge guides inference (priors), and how coherence is maintained as new observations are integrated (consistency). These dimensions are conceptually separable, enabling methods to be interpreted as combinations of design choices across axes, with each dimension representing a configurable aspect of the system. The outer branches illustrate representative examples for each category, without implying exclusivity.}
    \label{fig:constructing}
\end{figure*}

This section highlights the main techniques used to extract the nodes and edges defined in~\Cref{sec:whatis3dsg} from raw data and organize them into a \tdsgshort.
In the literature, the terminology for this task varies, reflecting the specific perspectives and assumptions of different research communities. 

In computer vision, this task is often referred to as \tdsgshort \emph{prediction}~\cite{liu2022explore, lv2024sgformer, wang2024weaklysupervised, koch2024open3dsg, ma2025sgsg, heo2025object} or \tdsgshort \emph{learning}~\cite{koch2024sgrec3d, zhang2024egosg, wald2020learning3d}, as it typically involves training a model to infer semantic and relational structure.
These methods usually operate on a reconstructed 3D scene representation (\eg dense \pcd) to produce a 3D Semantic Scene Graph.

In contrast, the robotics community, where live streams from onboard sensors such as LiDARs, \acp{IMU}, or RGB-D cameras are often processed, tends to use the terms \emph{construction} or \emph{generation}~\cite{hughes2022hydra, chen2025irs, chang2023hydramulti, tourani2025vsgraphs}. 
Here, the emphasis extends beyond inference to include real-time sensor processing, pose estimation, and the incremental construction of the graph hierarchy as the scene is observed. 
In this context, the term prediction is often reserved for extrapolating unobserved portions of the environment~\cite{Cadena_2016, placed2023survey}, rather than for inferring relationships within an already reconstructed scene.
Here we follow this naming convention and refer to the process of creating a \tdsgshort as \emph{construction}.

Existing approaches to \tdsgshort construction can be broadly categorized by how input data is processed: \emph{batch} (offline) methods~\cite{wald2020learning3d, koch2024open3dsg, werby2024hierarchical, werby2025keysg} assume all observations are available upfront and processed jointly, while \emph{incremental} (online) methods~\cite{hughes2022hydra, hughes2024foundations, gu2024conceptgraphs, linok2025beyondbarequeries} handle data sequentially as it is received from a sensor stream.
The batch setting eliminates challenges related to long-term consistency, allowing methods to focus primarily on accurately predicting scene entities (\Cref{sec:constructing-nodes}) and their relationships (\Cref{sec:constructing-relations}). 
The incremental setting requires the graph to be continuously updated as new observations arrive, which requires data association, handling dynamic entities and long-term changes, as well as mitigating sensor noise.
In addition, if the sensor pose is not known \textit{a priori}, further challenges arise, such as odometry drift and maintaining global spatial consistency, which are typical concerns of online \ac{SLAM} systems as discussed in~\Cref{sec:constructing-consistency}. 
Note that these processing modes are not mutually exclusive. 
For example, some methods construct nodes incrementally while inferring edges in a batch manner~\cite{gu2024conceptgraphs, linok2025beyondbarequeries}.
\Cref{fig:constructing} provides a conceptual decomposition of the design space for building a \tdsgshort.

\subsection{Building Nodes}
\label{sec:constructing-nodes}
Building a node consists of identifying the entity to be represented, selecting the subset of measurements that correspond to it, and estimating its attributes.

\boldparagraph{Object Nodes} 
A dominant strategy for generating object and object-part nodes in 3DSG construction is to lift 2D instance segmentation masks into a 3D representation. 
Given RGB-D observations, per-frame segmentation masks are back-projected into 3D using depth and camera poses and are subsequently fused across viewpoints.
Multi-view aggregation suppresses noisy detections while accumulating complementary evidence from different perspectives.
Recent work explores the use of geometric \acp{FM} for this \cite{maggio2026founditfoundationmodelfirsttaskdriven3d,kassab2026lexisgmonocular3dscene}.

Early approaches~\cite{armeni2019scenegraph, rosinol2021kimera, hughes2022hydra} follow this paradigm using closed-set 2D instance segmentation models such as Mask R-CNN~\cite{he2017mask}. 
The resulting masks are projected into 3D and associated across frames either through explicit object tracking~\cite{maggio2024clio, steinke2025curbosg} or by leveraging geometric and semantic cues such as spatial overlap or feature similarity~\cite{gu2024conceptgraphs}. 
Observations of the same instance are then consolidated using fusion strategies such as Bayesian updates~\cite{rosinol2021kimera}, majority voting~\cite{armeni2019scenegraph}, or adaptive aggregation mechanisms~\cite{linok2025beyondbarequeries}.

The advent of \acp{FM} enabled the development of open-vocabulary \tdsgshort pipelines~\cite{gu2024conceptgraphs, maggio2024clio, werby2025keysg, rotondi2025fungraph, deng2024opengraph}.
These methods first derive candidate object categories from images, either via taggers like RAM~\cite{zhang2024recognize} or by parsing object mentions from descriptions obtained by prompting a \ac{VLM} like GPT-4o~\cite{hurst2024gpt}.
The resulting label set is then supplied to open-vocabulary detectors, such as GroundingDINO~\cite{liu2024grounding} or YOLO-World~\cite{cheng2024yolo}, to predict bounding boxes, which are subsequently refined into segmentation masks using models such as SAM~\cite{kirillov2023segment}, from which multimodal descriptors \cite{radford2021learning} are extracted.

In batch \tdsgshort construction, multi-view consistency is typically abstracted by assuming a globally fused, class-agnostic segmented \pcd as input: instance masks are given, while semantic labels must be predicted. 
A common paradigm encodes instances with pre-trained 3D backbones, followed by graph-level reasoning. 
The work of \citet{wald2020learning3d} first proposed this strategy, employing PointNet~\cite{qi2017pointnet} encoders to extract 3D features from the masked \pcd, processing them with a \ac{GCN}, and decoding object and predicate labels through MLP heads. 
This formulation has inspired numerous extensions~\cite{zhang2021edge, sarkar2023sgaligner, koch2024lang3dsg, ma2025heterogeneousgraphlearning, miao2025scenegraphloc}.
Beyond architectural refinements, several works incorporate auxiliary priors during training to enhance semantic reasoning, including class prototypes~\cite{zhang2021knowledge}, external knowledge bases~\cite{qiu2023knowledge}, and dataset-level statistics~\cite{yeo2025statistical}.

Language-geometry alignment has emerged as a particularly effective strategy. 
By projecting \pcd features into, for example, the CLIP~\cite{radford2021learning} shared vision-language semantic space, these methods leverage large-scale pre-trained models to improve inference. 
In particular, SGFormer~\cite{lv2024sgformer} integrates CLIP text embeddings with PointNet features via cross-attention, VL-SAT~\cite{wang2023vlsat} regularizes predictions with a multimodal oracle, and Universal Scene Graphs~\cite{wu2025universal} unify multiple multimodal sources within a shared representation.

\boldparagraph{Higher-Level Nodes} 
Higher-level nodes represent entities beyond individual objects, often grouping them into coarser semantic or spatial units. 
As discussed in~\Cref{sec:whatis3dsg:modeling}, no canonical hierarchy exists, and their definition is largely task- and environment-dependent.

In structured indoor scenes, rooms and floors naturally serve as candidates for this abstraction. 
Room nodes can be obtained through geometry-driven parsing that detects structural boundaries \cite{bavle2022situational} and aggregates disjoint regions~\cite{armeni20163d,armeni2019scenegraph}, or by slicing the 3D Euclidean signed distance field below the ceiling and performing 2D connected-component analysis to delineate individual spaces~\cite{rosinol2021kimera}. 
Floors can be obtained from height distributions and by partitioning rooms through planar spatial segmentation~\cite{werby2024hierarchical}.
Early approaches cluster free space to form higher-level nodes, improving efficiency while forgoing explicit semantics~\cite{hughes2022hydra, hughes2024foundations}. 
Recent methods mitigate this limitation by enriching geometrically derived 3D primitives with open-vocabulary embeddings~\cite{maggio2024clio}. 

In urban environments, road topology can be extracted from agent trajectories by detecting inflection and intersection points that define lane-graph nodes, while the resulting segments form higher-level region nodes~\cite{deng2024opengraph, steinke2025curbosg, greve2024collaborative}.
Beyond road networks, higher-level nodes can also emerge from semantic abstraction, either by inferring spatial ontologies with \acp{LLM}~\cite{strader2024spatial} or by aggregating perceived regions and objects from large-scale maps into functional areas~\cite{Nyffeler_2025_ICCV, yang2025h3dsgplant, xie2024embodied}.
Another line of work follows terrain-driven segmentation, clustering regions according to traversability or physical terrain properties~\cite{samuelson2025terrainawaretaskdriven, samuelson2025terra}.

\boldparagraph{Node Attributes}
Additional node properties are typically computed to support the requirements of the target application, either computed directly from the available 3D geometry \cite{chang2025ashita, chang2023context, wu2021scenegraphfusion}, inferred from visual observations using \ac{VLM}-based prompting~\cite{engelbracht2024spotlight, gu2024conceptgraphs, rotondi2025fungraph}, or obtained as textual descriptions (captions) and feature embeddings generated from images using \acp{VLM} or pre-trained backbones~\cite{gu2024conceptgraphs, koch2024open3dsg, werby2025keysg}.

\boldparagraph{Open Challenges: Perception Cascades, Ambiguous Object Boundaries, and Deferred Semantics}
Despite significant recent progress, constructing robust and semantically rich nodes remains an open challenge.
Current pipelines rely on cascaded perception modules rather than end-to-end node inference, leading to early errors that propagate and result in duplicated, fragmented, or missing nodes. 
Node extraction also assumes objects are discrete and well-bounded, whereas real environments often exhibit ambiguous boundaries, articulated structures, and object-scene continua (\eg built-in furniture or clutter).
Finally, extracting all semantic attributes of an entity at graph construction time is often impractical, particularly in real-time systems. 
In many cases, the information that will ultimately be relevant is not known in advance. 
This motivates the development of hybrid representations that preserve rich underlying information in a compact implicit representation while maintaining only a limited set of explicit semantic attributes. 
Additional semantic details can then be retrieved and made explicit on demand, enabling both efficient graph construction and flexible downstream reasoning.

\subsection{Building Edges}
\label{sec:constructing-relations}
Constructing edges in a \tdsgshort requires identifying related pairs of nodes and assigning predicates that describe their relationships.
Existing methodologies differ in whether relations are extracted jointly with the nodes or inferred afterward, and in whether edges connect nodes within the same layer or across different layers of the hierarchy.

\boldparagraph{Intra-layer edges}
In batch \tdsgshort construction, early approaches predict object relations by representing them as special nodes, either by combining the \pcds of object pairs \cite{wald2020learning3d}, or by combining nodes with spatially close ones to define a local neighborhood \cite{wu2021scenegraphfusion}.
The resulting graph is then processed with a \ac{GCN} while preserving the orientation of edge context (\eg one object to the right of another) for predicate classification.
Building on this, \citet{zhang2021edge} represent relations explicitly as multi-dimensional edges and model their interaction with node features, while~\citet{ma2025heterogeneousgraphlearning} formulate the \tdsgshort as a heterogeneous graph and organize predicates into meta-categories.
Several works refine the feature extraction by including knowledge graph priors~\cite{feng2023spatial, qiu2023knowledge, zhang2021knowledge} or open-vocabulary distillation~\cite{koch2024sgrec3d, koch2024lang3dsg}.

When building a \tdsgshort from images, geometric heuristics such as spatial overlap between nodes can be used either to directly create edges~\cite{linok2025beyondbarequeries, behrens2025lost} or to initialize a dense graph~\cite{gu2024conceptgraphs, zhan2025freeqgraph}, which is then pruned before querying an LLM to assign the corresponding predicates.
For open-vocabulary relationship extraction, some works consolidate predictions from a \ac{VLM} across all images for each node pair~\cite{rotondi2025fungraph, rotondi2025social3dsg}, while others first select a subset of views where the nodes are most clearly visible before performing inference~\cite{koch2024open3dsg}.
While most works focus on object-level relations, some instead model higher-level structure, \eg by linking rooms through door observations and free-space connectivity \cite{rosinol2021kimera, hughes2022hydra} or by inferring room adjacency via shortest-path computations \cite{honerkamp2024language}.

\boldparagraph{Inter-layer edges} 
The construction of higher-level nodes (\Cref{sec:constructing-nodes}) often naturally induces edges between layers of the hierarchy. 
For instance, when a floor is partitioned into rooms, parent–child edges can be directly created between the floor and its rooms \cite{werby2024hierarchical, werby2025keysg}. 
Similarly, when object parts are included in the hierarchy, they can be linked to their parent objects using geometric cues~\cite{rotondi2025fungraph, jiang2024roboexp, wang2025curiousbot}. 
Beyond geometry, inter-layer associations can also be established through temporal signals such as scene changes \cite{engelbracht2024spotlight} or by leveraging commonsense knowledge from \acp{LLM}~\cite{zhang2025functional3dsg} to infer functional relations (\eg associating a light switch with the lamp it controls).

\boldparagraph{Open Challenges: Long-Range Relations, Cascade Failures and Scalability}
Edge construction remains challenging for three main reasons. 
First, many methods rely on geometric heuristics to select candidate node pairs, improving efficiency but often missing long-range or functional relations, particularly when only co-visible entities in the same image are considered.
Functional relationships are especially difficult to infer from appearance alone: observing a light switch, for instance, provides little evidence about which fixture it controls.
Second, establishing relations depends heavily on upstream node extraction, where segmentation errors, occlusions, or inconsistent identities propagate into incorrect edges. 
Third, scalability remains an open challenge, not only due to the computational cost of evaluating a large number of candidate pairs, but also due to the rapid growth in the number of potentially meaningful relationships as scene complexity increases. 
Enumerating and maintaining all possible relations is often neither practical nor desirable, as many relations may be weak, redundant, or only relevant in specific task contexts. 
This motivates approaches that represent relationships selectively and generate, retrieve, or refine them on demand rather than constructing a fully connected relational graph upfront.

\subsection{Maintaining Consistency}
\label{sec:constructing-consistency}
While \Cref{sec:constructing-nodes,sec:constructing-relations} discuss how individual posed observations are associated with existing nodes, incremental pipelines face two additional challenges depending on their input.
When operating without ground-truth pose, the method must maintain \textit{spatial consistency}, ensuring that all graph entities remain geometrically coherent with the underlying metric map as the robot's estimate of the world evolves.
Moreover, when processing observations in dynamic environments, the pipeline must additionally maintain \textit{temporal consistency}, ensuring that the graph remains valid as objects move, appear, or disappear over time.
We refer to pipelines that must address either of these challenges as \textit{SLAM-integrated} \tdsgsshort, as they must satisfy consistency requirements analogous to those in \ac{SLAM} systems~\cite{slam-handbook}.

\boldparagraph{Spatial consistency}
At the core of incremental spatial consistency lies data association. 
In practice, most methods ensure it by relying on strategies inherited from object \ac{SLAM}. 
Local detections are scored relative to existing map entities using geometric distance, overlap, appearance similarity, or embedding distance, followed by greedy or linear assignment~\cite{gu2024conceptgraphs,maggio2024clio,rosinol2021kimera,hughes2022hydra,hughes2024foundations}.
Pose priors can be estimated from underlying odometry systems \cite{rosinol2021kimera, hughes2022hydra, hughes2024foundations} and global consistency can be ensured by extracting descriptors~\cite{hughes2022hydra, hughes2024foundations} or keyframes~\cite{kim2025tacs} from the \tdsgshort hierarchy for loop closure.
The S-Graph family~\cite{bavle2022situational,bavle2023sgraphs,bavle2025sgraphs2,tourani2025vsgraphs} takes a complementary approach by maintaining the full \tdsgshort as an optimizable factor graph, so that structural entities such as walls, rooms, and floors directly constrain pose estimation rather than being constructed post-hoc. 
A key insight across this line of work is that floor-level constraints help disambiguate loop closures in multi-floor buildings, where naive global corrections would otherwise introduce false associations across different levels.
In multi-agent settings, spatial consistency becomes a shared-frame alignment problem: each robot builds its partial graph in its own drifting reference frame, requiring inter-robot loop closures and cross-agent data association to merge them without duplicating entities~\cite{chang2023hydramulti}. 
CURB-SG~\cite{greve2024collaborative} further shows that anchoring this association to persistent geometric structures, such as lanes and curbs in the driving context, is more reliable than associating through dynamic or semantically ambiguous entities.

\boldparagraph{Temporal consistency} 
The first step toward ensuring temporal consistency is to detect and track dynamic entities in the scene, ensuring consistency between consecutive observations.
This includes both identifying which entities are dynamic and maintaining their identity across frames.
Early approaches \cite{rosinol2021kimera, rosinol2020dynamic} assumed that only humans move within a scene, tracking them using SMPL~\cite{SMPL:2015} meshes. 
In contrast,~\citet{behrens2025lost} estimate the probability of hand–object interaction from egocentric observations to start the tracking.
More recent methods~\cite{steinke2025curbosg, gorlo2025describe, yan2025dynamicopenvocabulary, ferraina2026lost} explicitly distinguish and track dynamic objects.
These approaches address the short-term tracking problem of maintaining entity identity across consecutive frames.
Long-term consistency instead aims to ensure consistency across separate visits to a scene, often relying only on partial observations (see \cite{sh-ch15-dyndef} for a detailed discussion).
To this end, approaches typically rely on change detection to update the graph. \citet{looper2023vsg} propose a GCN-based architecture to learn the likelihood of entity motion, \citet{ge2025dynamicgsgd} detect changes using a Gaussian map, \citet{nguyen2025efficientattribute} learn to cluster and match object instances, and \citet{kabalar2023longtermretrieval} use appearance descriptors and category-based filters.

\boldparagraph{Open Challenges: Semantic Consistency, Forgetting, and Marginalization} 
Current approaches to maintaining consistency in SLAM-integrated \tdsgshort pipelines remain limited in several key aspects.
A primary shortcoming is that consistency is treated almost exclusively as a geometric problem: methods enforce spatial alignment across visits, but rarely reason about whether the Scene Graph remains \emph{semantically} consistent over time.
Whether an entity's open-vocabulary embedding, attributes, or relations agree across observations (and how uncertain those semantic estimates are) is rarely modeled explicitly, even though such reasoning could disambiguate data association and reveal changes that geometry alone cannot.
Moreover, reliable motion and change detection remains a challenge in its own right: detection accuracy is sensitive to parameter tuning and observation resolution, and scaling detectors to larger environments or finer resolutions quickly degrades either accuracy or tractability.
From a temporal perspective, existing research focuses predominantly on tracking the trajectories of dynamic objects rather than reasoning about the structural or relational shifts that such motion induces within the graph itself. 
Finally, principled mechanisms for managing the graph's information content over time remain open along two distinct axes.
The first is marginalization based on confidence: deciding when data association is certain enough, both spatially and temporally, to commit to a hard assignment, versus retaining a soft association that is more flexible but more expensive to store and reason over.
The second is compression and forgetting: abstracting away entities that resist stable instance-level representation, for example by replacing the explicit node of a frequently moving object with a descriptor or distribution attached to its enclosing room or place, or by decaying and eventually discarding unconfirmed entities.

\section{How are 3D Scene Graphs used and evaluated?}
\label{sec:applications}

\newcommand{\blockcell}{%
  \leftskip=0pt plus 1fill
  \rightskip=0pt plus -1fill
  \parfillskip=0pt plus 2fill
  \spaceskip=0.32em plus 0.55em minus 0.09em
  \tolerance=3000
  \emergencystretch=0pt
}
\colorlet{lanebg}{groupbg!20}
\newcommand{\lanepad}{\endgraf\vspace{2.4pt}}
\renewcommand{\tabularxcolumn}[1]{p{#1}}
\newcolumntype{P}{>{\columncolor{lanebg}[1.2pt][1.2pt]\hsize=1.10\hsize\blockcell\arraybackslash\strut}X<{\lanepad}}  %
\newcolumntype{C}{>{\columncolor{lanebg}[1.2pt][1.2pt]\hsize=1.05\hsize\blockcell\arraybackslash\strut}X<{\lanepad}}  %
\newcolumntype{S}{>{\columncolor{lanebg}[1.2pt][1.2pt]\hsize=1.15\hsize\blockcell\arraybackslash\strut}X<{\lanepad}}  %
\newcolumntype{L}{>{\columncolor{lanebg}[1.2pt][1.2pt]\hsize=1.00\hsize\blockcell\arraybackslash\strut}X<{\lanepad}}  %
\newcolumntype{M}{>{\columncolor{lanebg}[1.2pt][4pt]\hsize=0.70\hsize\blockcell\arraybackslash\strut}X<{\lanepad}}  %
 
\newcommand{\nocites}{\multicolumn{1}{>{\columncolor{lanebg}[1.2pt][1.2pt]}c}{\textcolor{black!30}{--}}}
\newcommand{\nocitesM}{\multicolumn{1}{>{\columncolor{lanebg}[1.2pt][4pt]}c}{\textcolor{black!30}{--}}}
 
\newcommand{\datasetitem}[1]{%
  \par\hangindent=1.6em\hangafter=1\hspace*{0.7em}#1%
}
\newcommand{\datasetcell}[1]{%
  \parbox[t]{\linewidth}{\raggedright\strut\vspace*{-\baselineskip}#1\strut}%
}
 
\newcommand{\grouprow}[1]{%
  \rowcolor{groupbg}%
  \multicolumn{6}{@{}l@{}}{\textcolor{groupfg}{\itshape\strut #1}\rule[-2.2pt]{0pt}{0pt}}\\
}
 
\makeatletter
\newcounter{ARfullwidthtable}
\newcommand{\ARfullwidthtablebox}[1]{%
  \refstepcounter{ARfullwidthtable}%
  \noindent
  \makebox[\textwidth][l]{%
    \ifcsname r@ARfullwidthtable:\theARfullwidthtable\endcsname
      \ifodd\getpagerefnumber{ARfullwidthtable:\theARfullwidthtable}\relax
      \else
        \hspace*{-\extramargin}%
      \fi
    \fi
    \parbox[t]{\typewidth}{#1}%
  }%
  \label{ARfullwidthtable:\theARfullwidthtable}%
}
\makeatother
 
{%
\makeatletter
\def\NAT@cmprs{\z@}
\makeatother
\let\oldcite\cite
\renewcommand{\cite}[1]{\citealp{#1}}
\newcommand{\dcite}[1]{\oldcite{#1}}
 
\begin{table*}[t]
    \caption[Paper classification by application and evaluation dataset.]%
        {Paper classification by application and evaluation dataset.
        }
    \label{tab:applications}
  \ARfullwidthtablebox{%
    \scriptsize
    \setlength{\tabcolsep}{4pt}
    \setlength{\extrarowheight}{2.2pt}
    \renewcommand{\arraystretch}{1.12}
 
\begin{tabularx}{\typewidth}{@{}>{\raggedright\arraybackslash}p{0.17\typewidth}PCSLM@{}}
  \toprule
  \textbf{Dataset}
    & \multicolumn{1}{c}{\shortstack[c]{\textbf{Point Cloud-Based}\\\textbf{Construction}}}
    & \multicolumn{1}{c}{\shortstack[c]{\textbf{Image-Based}\\\textbf{Construction}}}
    & \multicolumn{1}{c}{\shortstack[c]{\textbf{Scene}\\\textbf{Understanding}}}
    & \multicolumn{1}{c}{\shortstack[c]{\textbf{Planning \&}\\\textbf{Navigation}}}
    & \multicolumn{1}{c}{\textbf{Manipulation}} \\
  \midrule
 
  \grouprow{3RScan-based}
  \datasetcell{%
    \datasetitem{3RScan~\dcite{wald2019rio}}%
    \datasetitem{3DSSG~\dcite{wald2020learning3d}}}
    & \cite{wald2020learning3d,wu2021scenegraphfusion,qiu2023knowledge,feng2023spatial,wang2023vlsat,hou2025fross,xu2025tb,liu2022explore,koch2024open3dsg,chen2024clip,koch2024sgrec3d,zhang2024egosg,zhang2021knowledge,lv2024sgformer,wang2024weaklysupervised,koch2025relationfield,koch2024lang3dsg,heo2025object,ma2025heterogeneousgraphlearning,yeo2025statistical,zhan2025freeqgraph,wu2025universal,zhang2021edge}
    & \nocites
    & \cite{looper2023vsg,sarkar2023sgaligner,koch2025relationfield,ma2025sgsg,miao2025scenegraphloc,singh2025sgaligner++,kabalar2023longtermretrieval,chen2024scene}
    & \cite{bartoli2025longterm}
    & \nocitesM \\
    \grouprow{ARKitScenes-based}
  \datasetcell{%
    \datasetitem{ARKitScenes~\dcite{baruch2021arkitscenes}}%
    \datasetitem{SceneFun3D~\dcite{delitzas2024scenefun3d}}%
    \datasetitem{FunGraph3D~\dcite{zhang2025functional3dsg}}}
    & \nocites
    & \cite{werby2025keysg,rotondi2025fungraph,zhang2025functional3dsg,feng2026tfuns3dtask}
    & \cite{werby2025keysg,rotondi2025fungraph,zhang2025functional3dsg,feng2026tfuns3dtask}
    & \nocites
    & \cite{rotondi2025fungraph,zhang2025functional3dsg} \\
    \grouprow{Gibson-based}
  \datasetcell{%
    \datasetitem{Gibson~\dcite{xia2018gibson}}%
    \datasetitem{Armeni et al.~\dcite{armeni2019scenegraph}}%
    \datasetitem{iGibson~\dcite{shen2021igibson}}}
    & \nocites
    & \cite{armeni2019scenegraph,honerkamp2024language}
    & \nocites
    & \cite{agia2022taskography,rana2023sayplan,liu2025delta,honerkamp2024language,musumeci2025context,dai2024planning}
    & \cite{honerkamp2024language} \\
    \grouprow{KITTI-based}
  \datasetcell{%
    \datasetitem{KITTI~\dcite{geiger2013vision}}%
    \datasetitem{SemanticKITTI~\dcite{behley2019semantickitti}}}
    & \nocites
    & \cite{deng2024opengraph,stathoulopoulos2025havewe}
    & \cite{deng2024opengraph}
    & \cite{ray2024taskmotion}
    & \nocitesM \\
    \grouprow{MP3D-based}
  \datasetcell{%
    \datasetitem{Matterport3D~\dcite{chang2017matterport3d}}%
    \datasetitem{HM3D~\dcite{ramakrishnan2021habitat}}%
    \datasetitem{HM-EQA~\dcite{ren2024explore}}}
    & \nocites
    & \cite{hughes2024foundations,werby2024hierarchical,strader2024spatial,linok2025indoorgrounding,chang2025ashita,chen2025irs}
    & \cite{xu2025tb,werby2025keysg,linok2025indoorgrounding,saxena2024grapheqa,strader2024spatial,saucedo2024beliefsg,saucedo2025estimatingcommonsense,kassab2026lexisgmonocular3dscene}
    & \cite{werby2024hierarchical,chang2025ashita,saxena2024grapheqa,loo2024openscenegraphsopen,yin2024sg,yin2025unigoal,hu2025imaginative}
    & \nocitesM \\
    \grouprow{Replica-based}
  \datasetcell{%
    \datasetitem{Replica~\dcite{straub2019replica}}%
    \datasetitem{ReplicaSSG~\dcite{hou2025fross}}}
    & \cite{hou2025fross}
    & \cite{gu2024conceptgraphs,maggio2024clio,ge2025dynamicgsgd}
    & \cite{gu2024conceptgraphs,wang2025gaussiangraph,maggio2024clio,werby2024hierarchical,ge2025dynamicgsgd,werby2025keysg,linok2025beyondbarequeries,zhan2025freeqgraph}
    & \nocites
    & \nocitesM \\
    \grouprow{ScanNet-based}
  \datasetcell{%
    \datasetitem{ScanNet~\dcite{dai2017scannet}}%
    \datasetitem{ScanNet++~\dcite{yeshwanth2023scannet++}}%
    \datasetitem{ScanRefer~\dcite{chen2020scanrefer}}%
    \datasetitem{Sr3D/Nr3D~\dcite{achlioptas2020referit_3d}}}
    & \cite{kim2019sparse3d,wu2021scenegraphfusion,koch2024open3dsg}
    & \cite{tourani2025vsgraphs,ge2025dynamicgsgd}
    & \cite{linok2025indoorgrounding,xu2025tb,wang2025gaussiangraph,werby2024hierarchical,werby2025keysg,tourani2025vsgraphs,linok2025beyondbarequeries,zemskova20253dgraphllm,miao2025scenegraphloc,zhan2025freeqgraph,chang2023context,singh2025sgaligner++}
    & \cite{kim2019sparse3d,xu2024point2graph}
    & \nocitesM \\
    \grouprow{Simulator-based}
  \datasetcell{%
    \datasetitem{AI2-THOR~\dcite{kolve2017ai2}}%
    \datasetitem{RoboTHOR~\dcite{deitke2020robothor}}%
    \datasetitem{ProcTHOR~\dcite{deitke2022procthor}}%
    \datasetitem{CARLA~\dcite{Dosovitskiy17}}}
    & \nocites
    & \cite{greve2024collaborative}
    & \nocites
    & \cite{yin2024sg,yin2025unigoal,rajvanshi2024saynav,rivera2024conceptagent,talukder2024anticipatoryplanning,rajvanshi2025sayconav}
    & \nocitesM \\
    \grouprow{Custom}
  \datasetcell{\datasetitem{(various)}}
    & \cite{ozsoy2024holistic}
    & \cite{hughes2022hydra,samuelson2025terrainawaretaskdriven,millanromera2025metricsemanticfactorgraph,samuelson2025terra,hughes2024foundations,maggio2024clio,chang2023hydramulti,tourani2025vsgraphs,bavle2022situational,bavle2023sgraphs,gorlo2025describe,buechner2026momasg,yang2025h3dsgplant,bavle2025sgraphs2,kim2025tacs,strader2025hierarchicalplanning,ribeiro2025collaborativeslam}
    & \cite{ozsoy2024holistic,strader2024spatial,xie2024embodied,nguyen2025efficientattribute,gorlo2025describe,tourani2025vsgraphs,engelbracht2024spotlight,gu2025artisgfunctional3dscene,ma2025sgsg,gorlo2024trajectory,chang2023context,behrens2025lost,rotondi2025social3dsg,buechner2026momasg,ferraina2026lost,jiang2025exploring3dactivity,cheng2025spatialrgpt,ray2025structuredinterfacesautomatedreasoning,maggio2026founditfoundationmodelfirsttaskdriven3d}
    & \cite{bavle2022situational,yan2025dynamicopenvocabulary,ravichandran2022hierarchical,werby2024hierarchical,wang2025curiousbot,viswanathan2024xflieleveragingactionablehierarchical,chen2025irs,gu2024conceptgraphs,viswanathan2025spade,samuelson2025terra,rana2023sayplan,honerkamp2024language,strader2025hierarchicalplanning,ray2024taskmotion,shan2025graph2nav,jiang2025exploring3dactivity,qi2025composebyfocus,ray2025structuredinterfacesautomatedreasoning}
    & \cite{wang2025curiousbot,gu2024conceptgraphs,gu2025artisgfunctional3dscene,engelbracht2024spotlight,yan2025dynamicopenvocabulary,chang2023context,buechner2026momasg,jiang2024roboexp,honerkamp2024language,rivera2024conceptagent,yu2025articulated3Dsg,qi2025composebyfocus} \\
 
  \bottomrule
\end{tabularx}%
\par\vspace{3pt}
{\scriptsize\itshape We recommend the
\href{https://chromewebstore.google.com/detail/google-scholar-pdf-reader/dahenjhkoodjbpjheillcadbppiidmhp?hl=en}{Google Scholar PDF Reader}
for browsing the table.}
  }
\end{table*}
}

The value of a \tdsglong lies not only in how accurately it represents a scene, but also in how effectively its structure can support downstream applications. 
In practice, however, these two aspects often remain only partially aligned: many works construct rich graph representations, yet downstream tasks frequently use only a limited subset of their semantic, spatial, or functional information. 
This section, therefore, reviews \tdsgsshort from two complementary perspectives. 
We first discuss how the intrinsic quality of the representation is evaluated. 
We then examine how \tdsgsshort are used across their main application domains, highlighting both their current benefits and the limitations that prevent applications from fully leveraging their representational richness. 
This perspective also helps identify open challenges and future directions for improving the effectiveness of \tdsgsshort as general-purpose environment models. 
\Cref{tab:applications} further classifies the reviewed works by the datasets used to evaluate specific tasks.

\subsection{Intrinsic Evaluation}
\label{sec:applications-eval}

Evaluating the intrinsic quality of a \tdsgshort primarily concerns the correctness of its nodes and edges.
Since most pipelines derive graph nodes directly from segmented objects or regions, node evaluation is often closely related to instance-level semantic segmentation.
Accordingly, node quality is typically measured using standard detection metrics such as \emph{precision}, \emph{recall}, and \emph{F1-Score}, segmentation metrics such as \emph{mean Intersection-over-Union} (mIoU) and \emph{average precision}, as well as 3D reconstruction metrics such as \emph{Chamfer distance} across the most popular 3D datasets~\cite{wald2019rio, straub2019replica, dai2017scannet, chang2017matterport3d, baruch2021arkitscenes}. 
In the open-set case, evaluation is complicated by additional factors, such as label similarity and ambiguity, which are commonly assessed using metrics such as \emph{set ranking} or \emph{top-N frequency at category}~\cite{kassab2025openlex3d}.

Evaluating non-object nodes and relationships between nodes is even more challenging because they are often ambiguous and task-dependent, making it difficult to define a single ground-truth structure. 
Moreover, even when a ground-truth graph is available, computing graph similarity remains NP-complete; therefore, directly comparing the predicted graph to it via graph edit distance is computationally expensive. 
To address this, relationships are often evaluated using triplet-based metrics~\cite{wald2020learning3d} such as 
\emph{recall@k}, \emph{predicate classification accuracy}, or \emph{triplet prediction accuracy}. 
In practice, only a limited number of datasets provide reliable ground-truth \tdsgshort annotations~\cite{armeni2019scenegraph, wald2020learning3d, hou2025fross, zhang2025irefvla, xu2025tb}, enabling quantitative evaluation of both nodes and edges.
Works evaluated on the 3DSSG dataset operate on instance-segmented but semantically agnostic \pcds in batch, whereas online methods on RGB-D often require custom datasets~\cite{maggio2024clio, samuelson2025terrainawaretaskdriven} or crowdsourced protocols \cite{gu2024conceptgraphs, rotondi2025social3dsg} for edge evaluation, especially for hierarchical graphs. 
We make this distinction explicit in \Cref{tab:applications}.
Finally, SLAM-integrated 3DSGs~\cite{bavle2022situational, hughes2022hydra, hughes2024foundations, kim2025tacs} also directly measure the accuracy of the resulting trajectory and map using localization metrics such as \emph{absolute trajectory error} (ATE).

\subsection{Scene Understanding}
\label{sec:applications-sun}
Scene understanding is often viewed as a primary motivation for building \tdsgsshort, as we want to construct them as a representation of our understanding of the scene.
However, in this section, we use the term scene understanding more specifically to refer to how \tdsgsshort support reasoning over the entities and relational structure of a scene, enabling tasks such as cross-modal entity grounding~\cite{wald2020learning3d, chen2024scene, miao2025scenegraphloc} and question answering~\cite{saxena2024grapheqa, gu2024conceptgraphs, linok2025beyondbarequeries}, with particular emphasis on their integration with \acp{LLM}~\cite{xie2024embodied, werby2025keysg, ray2025structuredinterfacesautomatedreasoning, zemskova20253dgraphllm, gorlo2025describe}.

\citet{wald2020learning3d} demonstrated that \tdsgsshort
enable cross-domain retrieval by comparing structural similarity across images, \pcds, language descriptions, and meshes on the 3RScan dataset~\cite{wald2019rio}.
Subsequent works~\cite{chen2024scene, miao2025scenegraphloc, zemskova20253dgraphllm} extended this idea by learning joint embedding spaces between language features and \tdsgshort representations.

To answer unstructured language queries such as ``the red chair next to the window'', early approaches compared CLIP features extracted from the query with those associated with graph nodes \cite{werby2024hierarchical, maggio2024clio}. 
While effective for simple referential queries, this strategy becomes limited when queries require resolving multiple pieces of information across different layers of the \tdsgshort hierarchy, for example ``how can I turn on the light in the bathroom on the second floor?''
To address this, recent approaches~\cite{rotondi2025fungraph, werby2025keysg} leverage a key advantage of \tdsgsshort first exploited by~\cite{rana2023sayplan}: their ability to be serialized into structured, \ac{LLM}-readable formats such as JSON. 
This representation allows general-purpose \acp{LLM} to reason over the hierarchical scene structure and answer complex multi-hop queries by operating on serialized nodes and edges~\cite{zhang2025functional3dsg, rotondi2025fungraph, saxena2024grapheqa}, which can include any additional semantic frame descriptions \cite{navigli-etal-2024-nounatlas} and properties~\cite{rotondi2025social3dsg, bartoli2025longterm}.
However, as the graph grows, the serialized JSON representation expands accordingly, quickly exhausting the \ac{LLM}'s context window. 
To address this limitation, subsequent works propose retrieving smaller, relevant subgraphs using heuristic strategies \cite{linok2025beyondbarequeries}, constructing the representation in a RAG-inspired manner to enable fast multimodal indexing~\cite{werby2025keysg}, and applying GraphRAG techniques~\cite{xie2024embodied, ray2025structuredinterfacesautomatedreasoning} to reduce token usage.
Less common approaches to interfacing with \acp{LLM} include providing detailed natural language descriptions of the scene derived from the \tdsgshort to predict scene evolution~\cite{gorlo2024trajectory}, and developing \ac{LLM}-based agents that directly interact with the \tdsgshort to answer spatio-temporal queries \cite{gorlo2025describe}.
More broadly, \citet{ray2025structuredinterfacesautomatedreasoning} propose exposing \tdsgsshort through a structured query interface, enabling \acp{LLM} to access and reason over scene information via tool-augmented agentic workflows.

\boldparagraph{Evaluation}
The objective of the methods described in this section is to retrieve the scene elements that best match a query~\cite{wald2020learning3d, chen2024scene, miao2025scenegraphloc}. 
For question answering and object grounding, evaluation is typically stratified by hierarchical reasoning levels, spanning from tasks focused on individual objects to those requiring scene-level or floor-level understanding~\cite{werby2024hierarchical, werby2025keysg}. 
It also accounts for query complexity, for instance distinguishing spatial and attribute-based reasoning~\cite{linok2025beyondbarequeries, saxena2024grapheqa, rotondi2025social3dsg}. Common benchmarks for these tasks are Sr3D~\cite{achlioptas2020referit_3d} and Nr3D~\cite{achlioptas2020referit_3d}, which provide natural language referring expression annotations grounded in the ScanNet~\cite{dai2017scannet} dataset.
Performance is commonly measured using \emph{recall@k} or \emph{accuracy@k} for entity grounding, based on semantic similarity or geometric overlap. 
In spatio-temporal question answering, performance is commonly measured using \emph{question accuracy}, together with \emph{positional} and \emph{temporal error} metrics that assess the spatial and temporal grounding of the predicted answer~\cite{anwar2025remembr,gorlo2025describe}.

\boldparagraph{Future Directions: Reasoning Benchmarks, Model Distillation and Query-Adaptive Graphs}
An important future direction is the development of richer benchmarks that evaluate long-horizon spatio-temporal reasoning, capture the nuances of multi-hop graph reasoning, and assess the faithfulness of model outputs to the underlying graph structure, particularly regarding how different edge-construction strategies influence downstream behavior.
Another promising direction is to reduce reliance on online \acp{LLM} by distilling the capabilities of general-purpose models into compact, scene-specific agents that can operate on embodied platforms under realistic compute and latency constraints. 
Equally important is the design of adaptive \tdsgsshort that can specialize to different queries, emphasizing the entities, relations, and temporal context most relevant to the task, thereby improving retrieval quality and downstream reasoning performance.

\subsection{Planning and Navigation}
\label{sec:applications-planning}
A central question in robotics is how to synthesize the sequence of actions a robot must execute to accomplish a given task.
Classical symbolic task planning assumes a task or goal specified as a set of predicates over discrete variables, typically representing objects or symbolic entities; this assumption is shared by both planning domain definition language (PDDL)~\cite{ghallab1998pddl} and linear temporal logic (LTL)~\cite{pnueli1977temporal} formulations.
Task planning was envisioned as an early application of \tdsgsshort: one of the first \tdsgshort works~\cite{kim2019sparse3d} demonstrated object-level task planning by grounding PDDL goals in the graph through hand-designed rules.
While subsequent research concentrated primarily on the construction of \tdsgsshort, \citet{agia2022taskography} initiated a line of work on \emph{scalable} task planning over \tdsgsshort, a challenge that many works~\cite{rana2023sayplan,liu2025delta,talukder2024anticipatoryplanning,chang2025ashita,musumeci2025context} have since taken up.
Proposed strategies range from formally pruning the search space~\cite{agia2022taskography,ray2024taskmotion}, to learning relevancy and cost priors~\cite{talukder2024anticipatoryplanning}, to exploiting pre-trained priors~\cite{rana2023sayplan,liu2025delta,chang2025ashita,musumeci2025context}.

A complementary line of work investigates how tasks expressed in natural language can be grounded to elements of a \tdsgshort.
\citet{rana2023sayplan} first demonstrated joint grounding and planning of natural language tasks using \acp{LLM}, followed by several others~\cite{rivera2024conceptagent,yan2025dynamicopenvocabulary,honerkamp2024language,werby2024hierarchical}.
Such approaches perform well when tasks fall within the training distribution of the underlying \acp{LLM}, but typically sacrifice formal planning guarantees such as completeness and soundness.
To retain these guarantees, other approaches~\cite{strader2025hierarchicalplanning,dai2024planning,liu2025delta,musumeci2025context} instead use \acp{LLM} to translate natural language instructions into symbolic goals expressed in PDDL or LTL, and then solve the grounded problem with off-the-shelf planners.
Planning problems may further involve continuous variables, such as robot configurations, particularly for complex manipulation tasks; to date, only a single approach~\cite{ray2024taskmotion} addresses joint \ac{TAMP} on \tdsgsshort.
For natural language tasks, only \citet{buechner2026momasg} construct kinematic graphs and demonstrate planning over kinematic states using \acp{LLM}.

\boldparagraph{Robot Navigation and Exploration}
Since most planning work addresses reasoning and navigation tasks, we examine robot navigation with \tdsgsshort in more detail.
Robotic navigation requires identifying a collision-free path between robot configurations.
This, in turn, requires a notion of traversability, which often coincides with free space and is encoded through either explicit or implicit representations, such as probabilistic roadmaps or potential functions.
While safety remains paramount, navigation solutions increasingly incorporate social and semantic considerations.
Existing approaches differ in how they couple navigation with the \tdsgshort.
One line of work embeds navigation directly into the graph's hierarchical structure.
An early example, SayPlan~\cite{rana2023sayplan}, first performs iterative semantic search over the graph to produce a high-level plan, which is then refined into an executable path using Dijkstra's algorithm.
Follow-up works~\cite{werby2024hierarchical,honerkamp2024language,xu2024point2graph} extract free space from dense semantic maps and compute a Generalized Voronoi Diagram (GVD) to build sparse navigational graphs that are interleaved with the \tdsgshort hierarchy. 
A second set of approaches~\cite{rajvanshi2024saynav, shan2025graph2nav, yan2025dynamicopenvocabulary, gu2024conceptgraphs, viswanathan2025spade} couples \tdsgsshort with metric maps by projecting object positions onto dense maps for cost-based point navigation.
A third line of work constructs \tdsgsshort in purely topological domains, \ie without depending on metric maps: goals are selected via \ac{LLM} reasoning over the graph hierarchy and reached via shortest-path search over its connectivity~\cite{loo2024openscenegraphsopen,xie2024embodied}.
To address robot exploration in partially observed environments, several works investigate frontiers using \tdsgsshort, incorporating the semantics of nearby objects~\cite{honerkamp2024language}, applying chain-of-thought prompting for semantic enrichment~\cite{yin2024sg}, encoding goal conditions as local \tdsgsshort for matching~\cite{yin2025unigoal}, and extending this paradigm through predictive world modeling~\cite{hu2025imaginative}.

\noindent\boldparagraph{Evaluation} 
Task planning approaches using \tdsgsshort are concerned with downstream metrics relating to planner speed (\eg \emph{solver time}~\cite{liu2025delta, musumeci2025context} or \emph{planner iterations}~\cite{agia2022taskography}), plan quality (\eg \emph{success rate}~\cite{musumeci2025context}, \emph{plan length} and \emph{plan optimality}~\cite{dai2024planning}), or problem complexity (\eg \emph{size of state space}~\cite{liu2025delta}).
Evaluation of these works tends to rely on the dataset of~\citet{armeni2019scenegraph}, whose rich semantic annotations facilitate conversion into PDDL representations~\cite{agia2022taskography}.
Metrics such as \textit{success rate} extend to navigation approaches as well, though specialized metrics that examine path quality like \textit{success weighted by path length} (SPL)~\cite{yin2025unigoal, yin2024sg, rajvanshi2024saynav} and \textit{search efficiency}~\cite{honerkamp2024language} are common.
These approaches are predominantly validated in simulation, leveraging environments tightly coupled with datasets such as Habitat~\cite{ramakrishnan2021habitat}, AI2-THOR~\cite{kolve2017ai2}, or iGibson~\cite{shen2021igibson}.
Evaluation on real robot platforms, while desirable, is limited; this is due in part to the lack of capabilities or the inherent limits of robots that prevent a rich investigation of planner capabilities, and in part to static or full-observability assumptions in approaches, which make statistically significant trials expensive.

\noindent\boldparagraph{Future Directions: Dynamics, Scene Fidelity, and Tighter Integration}
Their sparse and symbolic nature makes \tdsgsshort particularly amenable to interaction with \acp{LLM}.
However, existing approaches remain largely confined to static, fully observed scenes and to comparatively simple tasks and planning domains (\eg mobile pick-and-place), which points to several main future directions.
One of these is the exploration of \tdsgsshort as a planning representation for highly dynamic or partially observed environments~\cite{wang2025curiousbot}, thereby lowering the barrier to demonstrations on real robot platforms.
Another future direction is to improve the fidelity of the \tdsgsshort{} in ways that are useful to the planner to enable higher-quality or more complex plans; this may involve modeling information such as kinematic states~\cite{buechner2026momasg} or exposing and exploiting material or semantic properties (\eg{} ``fragile'', ``expensive'') or affordances of objects to the planner.
Finally, future work should pursue a tighter coupling between planners and the \tdsgshort, for instance by modeling the robot's interactions with the scene or by letting the graph serve more explicitly as the planner's memory.

\subsection{Manipulation}
\label{sec:applications-manip}
In manipulation, \tdsgslong play a dual role: they provide the semantic and spatial grounding that manipulation tasks require, while manipulation itself can serve to build and enrich the graph.
Focusing on free-form text queries for pick-and-place, \citet{chang2023context} address target ambiguity by exploiting the spatial relationships and node features of the graph.
The robot gripper is controlled using the bounding box of the object to be retrieved to complete the task.
In general, \tdsgsshort support manipulation by providing semantic grounding~\cite{gu2024conceptgraphs, yan2025dynamicopenvocabulary}, integrating with grasping foundation models~\cite{fang2023anygrasp}, and enabling action reasoning for interactive search~\cite{honerkamp2024language, rivera2024conceptagent}.

In contrast, \citet{jiang2024roboexp} treat manipulation not only as a downstream task but also as a means for actively acquiring additional information about the environment. 
In their tabletop scenario, the robot iteratively solves a \ac{POMDP} to determine which interactions will best reveal new information about the scene, such as opening a drawer to inspect its contents. After exploration, the scene is autonomously reset to its original configuration, and all actions required to reveal a specific object are stored in the graph: for instance, retrieving a fork inside a cabinet may require first moving a condiment bottle that blocks the door and then opening the cabinet.
Building on this idea, \citet{wang2025curiousbot} propose a method for mobile robots operating in room-scale environments. 
The system is equipped with a fixed set of low-level skills such as ``open'', ``lift'', and ``push'', and its objective is to minimize unknown space by discovering as many nodes as possible using these skills.
A major limitation of this approach is that the skills are tailored to a single robotic platform; whenever a different robot is used, new skills must be created and arbitrarily selected for the specific scene.
To mitigate this limitation, \citet{qi2025composebyfocus} investigate how atomic skills can be learned directly from a \tdsgshort-based representation. 
Their results show not only that this is feasible, but also that learning from \tdsgshort outperforms other baselines in terms of policy robustness and generalization, owing to its semantic graph structure, a result consistent with the findings of \citet{cheng2025spatialrgpt}.

In particular, the richer and more detailed a \tdsgshort becomes, the more useful it can be for downstream tasks.
In line with this, \citet{rotondi2025fungraph} and \citet{zhang2025functional3dsg} show that adding a layer of functional interactive elements such as ``knobs'', ``handles'', and ``buttons'' improves manipulation by enabling robots to grasp the correct parts of the object more effectively. Approaching the problem from a one-shot distillation perspective, \citet{yu2025articulated3Dsg}, \citet{buechner2026momasg}, and \citet{gu2025artisgfunctional3dscene} use human demonstrations to recover articulation models for these parts, which are then used for accurate manipulation.

\boldparagraph{Evaluation} 
\tdsgsshort and manipulation capabilities are advancing in tandem with the release of curated datasets that include both fine-grained functional element annotations~\cite{delitzas2024scenefun3d} and \tdsgsshort augmented with functional nodes~\cite{zhang2025functional3dsg}.
However, their evaluation frameworks remain largely disconnected.
The standard evaluation metric across the literature~\cite{chang2023context, wang2025curiousbot, honerkamp2024language, yan2025dynamicopenvocabulary} is the \emph{success rate}, or \emph{step-wise success rate} for complex long-horizon tasks~\cite{yan2025dynamicopenvocabulary, honerkamp2024language, qi2025composebyfocus, rivera2024conceptagent}. 
In these evaluations, the role of the \tdsglong is primarily to localize the object to be manipulated within the scene, sometimes treating object detection as part of the manipulation pipeline itself~\cite{rivera2024conceptagent}.
\boldparagraph{Future Directions: Memory, Verifiability, and VLA Integration}
The limited role of the \tdsgshort at execution time leaves much of its accumulated knowledge unused, suggesting several extensions.
In particular, future systems could make more direct use of the graph's persistent representation of the environment, including articulation constraints, affordances, and state history.
Moreover, because the graph provides an explicit account of the scene, the preconditions and effects of actions can be checked against it.
This is most relevant for \acp{VLA}~\cite{kawaharazuka2025vla-survey}, which dominate current manipulation research yet condition only on the current observation and a language instruction.
Consider a metal bottle filled with water in a previous interaction: although visually indistinguishable from an empty one, its stored state ``full'' allows the policy to anticipate the added mass, adapt its grasp force, and avoid tilting during transport.
Beyond memory, grounding language instructions in graph nodes offers a path to improving \ac{VLA} generalization: rather than implicitly resolving ``the bottle on the shelf'' from pixels, the policy receives an explicit, verifiable binding between instruction and scene.
How best to expose a \tdsgshort to such models, whether as serialized context, retrieved subgraphs, or learned embeddings, remains an open question.

\subsection{Emerging Applications}
Beyond the most established use cases, recent works explore the integration of \tdsgslong with \ac{XR} systems for interactive scene refinement~\cite{ma2025sgsg, ribeiro2025collaborativeslam}, collaborative and multi-agent graph construction~\cite{chang2023hydramulti, rajvanshi2025sayconav, greve2024collaborative}, and graph compression and synchronization for distributed robotic systems~\cite{chang2023dlite, stathoulopoulos2025havewe}. 
\tdsgslong are also beginning to support \acp{FM} and policy learning by providing structured spatial context and relational abstractions for embodied agents~\cite{qi2025composebyfocus, cheng2025spatialrgpt}. 
Although many of these directions remain relatively underexplored and are often evaluated on specialized benchmarks, they already demonstrate that \tdsgslong can support substantially broader functionality than classical semantic mapping systems.

Another emerging trend is the inclusion of humans and social interactions directly within the graph representation. 
Early work modeled humans mainly through spatio-temporal links to objects and places~\cite{rosinol2020dynamic, rosinol2021kimera}, while more recent approaches, such as Social \tdsgslong~\cite{rotondi2025social3dsg}, incorporate human attributes, activities, and human-human as well as human-object relations in open-vocabulary settings. 
Structured human-centric representations have also shown value in specialized environments such as operating rooms~\cite{ozsoy2024holistic}.
Extending these ideas with language-based priors, such as \emph{``my desk is the one by the door of room 601''}, and persistent person-specific information, including preferences and routines, could improve situated \ac{HRI} and socially aware navigation.

Persistent and lifelong mapping, rather than static scene snapshots, could significantly expand the capabilities of future \tdsgsshort.
Initial efforts already maintain scene histories~\cite{schmid2024khronos, gorlo2025describe} and predict future object states~\cite{looper2023vsg}, but an important missing component is a predictive model of agent-environment dynamics that captures how embodied agents evolve through actions and interactions. 
The grounded and relational structure of \tdsgslong could naturally support scene editing~\cite{zhai2024echoscene}, scene generation \cite{Hu_2026_WACV}, and physically grounded simulation, enabling them to act as explicit world models enriched with semantic priors.

Beyond reasoning and prediction, the graph itself could serve as a mechanism for reconstruction, information sharing, and compression. 
Existing approaches discussed in this review reconstruct object instances independently, without explicitly leveraging similarities between observations of semantically or geometrically related entities.
A more effective approach would jointly reason over related instances, allowing geometry, semantic labels, and physical attributes to be estimated consistently within a unified parametric framework.
By recognizing repeated layouts, reusable templates, and recurring roles within the graph, such methods could improve reconstruction quality, support scene completion in partially observed environments, and enable compact representations of large-scale spaces such as buildings, campuses, and cities, where similar patterns naturally emerge.

\section{Conclusion}
\label{sec:conclusion}
Over the past six years, \tdsgslong have evolved from compact, offline descriptions of static scenes into incrementally constructed representations that support dynamic environments, interface with language models, and increasingly serve as the perceptual backbone of embodied AI systems.
This trajectory has unfolded along the three axes that organize this survey.
On the \emph{representation} side, the same hierarchical formalism has been progressively enriched with descriptive attributes, dynamic state, functional affordances, articulated kinematics, and human-centric context.
In \emph{construction}, the field has shifted from batch inference over pre-reconstructed scenes to incremental online pipelines that tolerate pose drift and dynamic content, replacing closed-set classifiers with open-vocabulary perception and enforcing global coherence through \ac{SLAM}-grade consistency mechanisms.
In \emph{applications}, \acp{LLM} have emerged as natural consumers of the serializable graph structure, enabling a general-purpose interface across navigation, manipulation, planning, embodied question answering, and beyond.

Nevertheless, the literature surveyed here also exposes several persistent gaps.
\emph{Evaluation} is the most systemic issue. 
Most work measures graph quality and downstream task performance independently, often on different datasets.
As a result, the effect of perception errors on downstream behavior is rarely quantified directly. 
Even when both are evaluated jointly, the metrics used often obscure the extent to which the graph representation itself contributes.
\emph{Spatial consistency} is absent from most learning-based pipelines, since batch processing on reconstructed scenes sidesteps the problem entirely.
\emph{Indoor-outdoor unification} remains underexplored, with the two regimes still relying on incompatible, ad-hoc decompositions.
\emph{Temporal reasoning} is still treated only marginally: the vast majority of methods assume quasi-static scenes, scene-history reconstruction is just emerging, and the interplay between functional state changes and graph dynamics remains conceptual.

Looking ahead, several directions stand out as both consequential and achievable.
\emph{Adaptive, task-conditioned graphs} that selectively expand or compress detail based on the active task are essential for scalable deployment on resource-constrained platforms. 
These should be coupled with lightweight parametric nodes that share appearance, semantics, and physical priors across recurring instances.
\emph{Distillation} of general-purpose \acp{LLM} into compact, scene-specific reasoning agents could reduce both cost and latency. 
At the same time, \emph{hybrid representations} that ground explicit graph structure in implicit geometry may combine the interpretability and modifiability of \tdsgsshort with the dense scene modeling capabilities of implicit representations.
Persistent and lifelong \tdsgsshort should be enriched with predictive dynamics. 
In particular, they should capture not only what changes, but also how agent-environment interactions reshape the graph. This is a prerequisite for treating \tdsgsshort as explicit world models rather than static scene snapshots.

Together, these directions point toward a broader shift in the role of \tdsgsshort: from descriptive scene abstractions to persistent, task-aware models capable of supporting perception, memory, and action within a unified representation.
By formalizing hierarchical, dynamic, and functional \tdsgsshort under a common definition, mapping the divide between batch and incremental construction, and analyzing evaluation practices across applications, this survey aims to provide the foundations on which the next generation of \tdsgsshort can be built: an actionable representation designed to support the future of robotics and physical intelligence.

\section{Author Contributions}
D.R. and F.A. co-led the project. D.R., F.A., S.K., N.H., M.B., and L.R.S. jointly conceptualized the work and defined the survey's structure and focus through weekly meetings over nine months, complemented by meetings with all authors.
D.R. wrote~\Cref{sec:intro,sec:related,sec:conclusion}, contributed to \Cref{sec:whatis3dsg,sec:constructing,sec:applications}, and developed the project website.
F.A. produced \Cref{tab:applications} and contributed to subsections of \Cref{sec:constructing,sec:applications}.
S.K. produced \Cref{fig:constructing}, helped in the creation of \Cref{fig:timeline} and contributed to subsections of \Cref{sec:applications}.
N.H. contributed to~\Cref{sec:whatis3dsg:modeling,sec:constructing-consistency,sec:applications-planning}.
M.B. produced \Cref{fig:timeline,fig:example3dsg} and contributed to \Cref{sec:applications-planning}.
L.R.S. wrote \Cref{sec:modelingdynamics} and provided extensive feedback and edits across all sections.
J.W., D.N., A.V., L.P., and F.T. provided feedback on the survey and suggested edits and corrections.
L.C. and K.O.A. provided extensive feedback on all drafts, suggested edits and corrections, and helped shape the final form of the survey.

\section{Acknowledgments}
This work has been supported by the German Federal Ministry of Research, Technology, and Space (BMFTR) under the Robotics Institute Germany (RIG) and the German Research Foundation Emmy Noether Program grant number 468878300.
The authors also thank the International Max Planck
Research School for Intelligent Systems (IMPRS-IS) for supporting Dennis Rotondi. This work has been carried out while Francesco Argenziano was enrolled in the Italian National Doctorate on Artificial Intelligence run by Sapienza University of Rome. 

{
    \small
    \bibliographystyle{ieeenat_fullname}
    \bibliography{references2}
}

\end{document}